\documentclass[journal]{IEEEtran}
\usepackage{mathrsfs}
\usepackage{amsmath}
\usepackage{amsfonts}
\usepackage{bbding}
\usepackage{amsmath,amsfonts}
\usepackage{algorithmic}
\usepackage{bm}
\usepackage{array}
\usepackage{caption}
\usepackage{textcomp}
\usepackage{stfloats}
\usepackage{url}
\usepackage{verbatim}
\usepackage{enumerate}
\usepackage{comment}
\usepackage{color}
\usepackage{algorithm}
\usepackage{cite}
\usepackage{makecell}
\usepackage{lipsum} %
\usepackage{multicol}

\usepackage{times}
\usepackage{epsfig}
\usepackage{amssymb}
\usepackage{subfigure}
\usepackage{multirow}
\usepackage{threeparttable}
\usepackage{arydshln}
\usepackage[colorlinks,urlcolor=black,linkcolor=black,anchorcolor=black,citecolor=black]{hyperref}
\newtheorem{Def}{\textbf{Definition}}
\newtheorem{Cor}{\textbf{Corollary}}

\newcommand{\e}{\text{e}}

\allowdisplaybreaks[2]
\def\BibTeX{{\rm B\kern-.05em{\sc i\kern-.025em b}\kern-.08em
    T\kern-.1667em\lower.7ex\hbox{E}\kern-.125emX}}
\usepackage{balance}

\begin{document}
\title{Rediscovering BCE Loss for Uniform Classification}
\author{Qiufu Li, Xi Jia, Jiancan Zhou, Linlin Shen, Jinming Duan
\thanks{Corresponding author: Linlin Shen.}
\thanks{Qiufu Li and Linlin Shen are with National Engineering Laboratory for Big Data System Computing Technology, Shenzhen University, Computer Vision Institute, Shenzhen University, 
and SZU Branch, Shenzhen Institute of Artificial Intelligence and Robotics for Society; Xi Jia and Jinming Duan are with School of Computer Science, University of Birmingham; Jiancan Zhou is with Aqara, Lumi United Technology Co., Ltd. }
\thanks{E-mail: qiufu\_li\_1988@163.com; x.jia.1@cs.bham.ac.uk; zhoujiancan@foxmail.com; llshen@szu.edu.cn; j.duan@bham.ac.uk}}

\markboth{March~2024: uniform classification}%
{Uniform Classification}

\maketitle

\begin{abstract}
This paper introduces the concept of uniform classification,
which employs a unified threshold to classify all samples rather than adaptive threshold classifying each individual sample.
We also propose the uniform classification accuracy as a metric to measure the model's performance in uniform classification.
Furthermore, begin with a naive loss, we mathematically derive a loss function suitable for the uniform classification,
which is the BCE function integrated with a unified bias.
We demonstrate the unified threshold could be learned via the bias.
The extensive experiments on six classification datasets and three feature extraction models show that,
compared to the SoftMax loss, the models trained with the BCE loss not only exhibit higher uniform classification accuracy but also higher sample-wise classification accuracy.
In addition, the learned bias from BCE loss is very close to the unified threshold used in the uniform classification.
The features extracted by the models trained with BCE loss not only possess uniformity
but also demonstrate better intra-class compactness and inter-class distinctiveness,
yielding superior performance on open-set tasks such as face recognition.
\end{abstract}


\section{Introduction}
Classification is one of the fundamental tasks in the field of machine learning.
For an $N$-class classification task, it typically involves comparing a sample to the $N$ class centers,
classifying the sample into the class whose center has an extremal metric (minimum distance or maximum similarity) with the sample.
We define the metric between the sample and the center of its actual class as its positive metric,
and the metrics between the sample and other class centers as its negative metrics.
Then the classification actually uses a threshold adaptive to the sample to distinguish its positive and negative metrics.
This type of classification is referred to as \textbf{point-wise} or \textbf{sample-wise classification}.
However, a unified threshold for distinguishing positive and negative metrics across all samples might be more useful in certain scenarios.
For example, in open-set tasks, if there exists such unified threshold,
it becomes straightforward to determine whether a new sample belongs to one of the $N$ known classes contained in the closed set, using this unified threshold.
The classification introduced here is referred to as \textbf{uniform classification}.

The uniform classification implies uniformity on the dataset (or its feature set),
which is a property different from both intra-class compactness and inter-class distinctiveness.
These three properties can be explained as follows:
intra-class compactness requires that each class of data be concentrated within a hypersphere centered at its class center with a specific radius,
and inter-class distinctiveness requires that the hyperspheres for each class do not contain data from other classes, meaning that the hyperspheres for each class do not intersect.
On the other hand, the uniformity requires that these non-intersecting hyperspheres have the same radius.

Classification generally does not directly compute the metrics between samples and class centers,
but rather calculate the metrics between their features.
This necessitates a good feature extraction model, where the loss function plays a crucial role in their training.
Currently, the most commonly used loss function in the classification is the cross-entropy/SoftMax loss, derived from probability statistics.
During the model training, for a given sample, the SoftMax loss maximizes the probability of classifying it into its true class,
obtaining an adaptive threshold for correct classification of that sample.
However, the SoftMax loss only imposes individual constraints on each sample and fails to impose a unified constraint on the all samples,
making it incapable of learning a unified threshold that can distinguish the positive and negative metrics across the all samples.
Besides the cross-entropy/Softmax loss, other commonly used losses for classification include
focal loss \cite{lin2017focal}, triplet loss \cite{schroff2015facenet}, center loss \cite{wen2016discriminative},
marginal SoftMax loss \cite{liu2016large}, and normalized SoftMax loss \cite{wang2017normface}, etc.
These losses similarly do not take into account the unified threshold.

In this paper, we introduce the concept of uniform classification and design its corresponding loss function.
We validate their effectiveness through extensive experiments.
The main contributions of this paper are summarized as follows.
\begin{enumerate}[~~(1)]
  \item We introduce, for the first time,
  the concepts of uniform classification and class-wise uniform classification.
  They involve a unified threshold to classify the all samples or, in the case of class-wise uniform classification,
  to classify the all samples of the same class using unified thresholds.

  \item We define the metrics such as sample-wise classification accuracy, class-wise uniform classification accuracy, and uniform classification accuracy
  tailored to the different types of classifications.
  These metrics serve to evaluate the performance of the feature extraction model and classifier in the different classification tasks.

  \item We have designed a loss function tailored for the uniform classification,
  which adopts the formula of binary cross-entropy (BCE) function but incorporates a unified bias.
  We have demonstrated that the bias could ultimately lead to the unified threshold after the training.

  \item In the experiments, we apply twelve losses based on SoftMax and BCE functions across six classification tasks, to train three  commonly used feature extraction models.
  The twelve losses utilize both linear classifier and its normalized variant, along with three different bias modes.
  The experimental results indicate that, compared to SoftMax loss, (a) BCE loss indeed has the capability to learn the unified threshold,
  making it more suitable for the uniform classification, (b) the models trained with the BCE loss extract features with better uniformity.
\end{enumerate}

The work is inspired by UniFace \cite{zhou2023uniface}, published at ICCV2023.
UniFace designs a unified threshold integrated cross-entropy (UCE) loss, and apply it to face recognition,
which achieves the best result in the MegaFace Recognition Challenge (MFR) at the time of its submission.
In comparison to UniFace and UCE, the present work further extends the idea of the unified threshold to general classification.
It introduces the concept of uniform classification, defines the corresponding metrics,
and validates through extensive experiments that BCE loss is superior to SoftMax loss in the uniform classification and even the general sample-wise classification.

\section{Related works}
\subsection{Uniformity}
Uniformity describes the homogeneous and uniform distribution of a property across the entirety of an object,
with no fundamental differences among its various parts.
Take the continuity of a function $f(x)$ in calculus for example.
The function $f(x)$ is continuous at a point $x$ if, for any given $\varepsilon>0$,
there exists a number $\delta>0$ (which might be related to $x$) such that $|f(x)-f(x_0)| < \varepsilon$ for $\forall~x_0$ with $|x-x_0|<\delta$.
On the other hand, the function is uniformly continuous if, for any given $\varepsilon>0$, there exists $\delta>0$ independent of any $x$,
such that $|f(x_0)-f(x_1)| < \varepsilon$ for $\forall x_0,x_1$ with $|x_0-x_1|<\delta$.

Similarly, one can define separability and uniform separability for a dataset.
For a dataset $\mathcal D$ comprising samples captured from $N$ categories,
its separability at a sample $\bm X$ implies that the sample can be correctly classified into its true category in some manner.
The classification typically involves determining $N$ class centers
and calculating $N$ metrics, such as distance or similarity, between the testing sample and the $N$ centers.
For instance, in a well-trained deep neural network used for image classification,
the rows of coefficient matrix of the last fully connected layer essentially learn the central features of each class.
For an image $\bm X$, after extracting its feature $\bm x$, calculate the inner product between $\bm x$ and each central feature,
then the image $\bm X$ is classified into the class corresponding to the maximum inner product.

If we define the metric between the sample $\bm X$ and its true category as its positive metric and that between it and other categories as its negative metrics,
then its positive metric must be an extremum among all metrics (the minimum distance or maximum similarity) to make it correctly classified.
In such a situation, there must exist a threshold $t_{\bm X}$ that can distinguish the sample's positive and negative metrics,
and then the dataset is separable at this sample.
Going one step further, if there exists a unified threshold $t$ that separates the positive and negative metrics of all samples,
then the dataset is uniformly separable.
These two different types of separability respectively evolve into the tasks of classification and uniform classification.

The existing literatures have employed the term of ``uniform classification'' \cite{yang2021validation,tizmaghz2022consistent},
while these works use a unified framework, system, or scheme, etc., to uniformly classify different objects with various properties.
In Banach space, there exists a theory of uniform classification \cite{benyamini1994uniform,benyamini2004introduction},
but we do not found any connection between it and the uniform classification proposed in this paper.

\subsection{Loss function}
As mentioned earlier, the classification typically does not directly measure the samples with the category centers,
but rather measure their features and the feature centers of each class,
which requires the elements such as loss functions to train well-performing feature extraction models.

Currently, the SoftMax or cross-entropy loss \cite{lecun1998gradient} is the most commonly used in the training of classification models.
In general, the feature $\bm x$ of a sample $\bm X$ is first transformed into $N$ metrics, $\{c_i(\bm x)\}_{i=1}^N$.
The loss then employs the SoftMax function to transform them into probabilities of the sample being predicted for each category,
 $\big\{\e^{c_i(\bm x)}/\sum_{j=1}^N \e^{c_j(\bm x)}\big\}_{i=1}^N$,
and finally calculate the cross-entropy with the probabilities $\{p_i(\bm x)\}_{i=1}^N$ of it truly belonging to each category,
i.e., $-\sum_i \big[p_i(\bm x)\cdot\log(\e^{c_i(\bm x)}/\sum_j \e^{c_j(\bm x)})\big]$.
When the sample is captured from the $i$-th category, only $p_i=1$, and all other $p_j$ are 0.
Thus, the SoftMax/cross-entropy loss is
\begin{align}
\label{eq_softmax_loss_initial}
\mathcal L_{\text{soft}}(\bm X) = -\log\frac{\exp\big(c_i(\bm x)\big)}{\sum_{j=1}^N\exp\big(c_j(\bm x)\big)}, ~\bm X \text{ from class } i.
\end{align}

The SoftMax loss function possesses many favorable properties, such as being continuously differentiable.
Depending on the practical classification problems, the SoftMax loss has been modified to create various other forms.
For examples, weighted SoftMax loss \cite{yue2017imbalanced} assign a weight to each class according to the proportion of samples of the class in the entire dataset,
to address the issue of unbalanced sample distribution in classes on the training set.
Normalized SoftMax loss \cite{wang2017normface} could enhance the stability of model training.
Marginal SoftMax loss \cite{liu2016large} introduces a margin between the positive and negative metrics of samples, enabling the model to learn features with better inter-class distinctiveness.
Focal loss \cite{lin2017focal} adjusts the confidence of samples belonging to each category,
tackling the problem of hard samples in the training set and strengthening the intra-class compactness of features.
The SoftMax loss is also combined with center loss \cite{wen2016discriminative} and triplet loss \cite{schroff2015facenet} to further enhance the intra-class compactness and inter-class distinctiveness of features.

Besides the intra-class compactness and inter-class distinctiveness, some tasks also require features to exhibit a certain consistency or uniformity.
According to the analysis and results presented in this paper, such consistency/uniformity cannot be achieved through the SoftMax loss,
but can be obtained through the binary cross-entropy (BCE) loss rediscovered in this study.
In fact, if the $N$-class classification task is regarded as $N$ binary classification tasks,
then the $j$-th metric $c_j(\bm x)$ of a sample $\bm X$ can be transformed into the probability of
whether it belongs to the category $j$ using the Sigmoid function, i.e., $\frac{\e^{c_j(\bm x)}}{1+\e^{c_j(\bm x)}}$.
The cross-entropy for the $j$-th binary classification task is $-\big[p_{j0}(\bm x)\cdot\log\frac{\e^{c_j(\bm x)}}{1+\e^{c_j(\bm x)}} + p_{j1}(\bm x)\cdot\log\frac{1}{1+\e^{c_j(\bm x)}}\big]$,
where $p_{j0}(\bm x)$ and $p_{j1}(\bm x)$ is the probabilities of the sample $\bm X$ belonging to or not to class $j$.
When $\bm X$ is captured from the $i$-th category, $p_{i0}=1, p_{i1}=0$, and $p_{j0}=0, p_{j1}=1$ for $j\neq i$,
then one can derive the BCE loss for the $N$-class classification,
\begin{align}
\label{eq_bce_loss_initial}
\mathcal L_{\text{bce}}(\bm X) = &\log\Big(1+\e^{-c_i(\bm x)}\Big) + \sum_{j=1\atop j\neq i}^N\log\Big(1+\e^{c_j(\bm x)}\Big).
\end{align}

In this paper, we will rederive the SoftMax loss and the BCE loss through mathematical calculation,
and uncover the connection between their biases and the feature uniformity.

\section{Classification}
\subsection{Dataset and feature set}
Suppose that $\mathcal{D}$ is a dataset captured from $N$ categories,
\begin{align}
\label{eq_sample_set}
\mathcal{D} = \bigcup_{i=1}^N\mathcal{D}_i
\end{align}
where $\mathcal{D}_i$ denotes the subset containing the samples from the same category $i$.
For each sample $\bm X$ in $\mathcal{D}$, a model $\mathcal {M}$ converts it into its feature $\bm x = \mathcal {M}(\bm X) \in \mathbb{R}^M$,
where $M$ is the length of the feature vector.
Then, for $\mathcal D$, one can get a feature set
\begin{align}
\label{eq_feature_set}
\mathcal{F} = \bigcup_{i=1}^N\mathcal{F}_i = \bigcup_{i=1}^N\big\{\bm x = \mathcal{M}(\bm X): \bm X\in\mathcal{D}_i\big\}.
\end{align}

A classifier $\mathcal {C}=\{c_i(\theta_i;\cdot)\}_{i=1}^N$ consists of $N$ real-value functions $c_i(\theta_i;\cdot),i=1,2,\cdots,N$,
mapping a sample's feature $\bm x = \mathcal {M}(\bm X)$ to $N$ classification metrics
\begin{align}
\label{eq_classification_metrics}
\big\{c_i(\bm x)=c_i(\theta_i;\bm x)\big\}_{i=1}^N\subset\mathbb{R},
\end{align}
where $\theta_i$ is the parameter set of the $i$-th metric function $c_i$.
The classifier $\mathcal {C}$ classifies the sample $\bm X$ into category $i$, if
\begin{align}
\label{eq_classification_condition_1}
c_i(\bm x) = \max\big\{c_j(\bm x)\big\}_{j=1}^N.
\end{align}

For any sample $\bm X$, we denote it as $\bm X^{(i)}$ if it was captured from category $i$, i.e., $\bm X^{(i)}\in\mathcal D_i$,
and we denote its feature as $\bm x^{(i)} = \mathcal M(\bm X^{(i)})$.
Then, using the classifier $\mathcal C=\{c_i(\theta_i;\cdot)\}_{i=1}^N$,
one can get its \textbf{positive classification metric} $c_i(\bm x^{(i)})$ and $N-1$ \textbf{negative classification metrics} $\{c_j(\bm x^{(i)})\}_{j=1\atop j\neq i}^N$.
The classifier $\mathcal C$ correctly classify the sample $\bm X^{(i)}$, if and only if
there exists a threshold $t_{\bm X^{(i)}}$ separating its positive and negative metrics, i.e.,
\begin{align}
\label{eq_classification_condition_2}
c_i(\bm x^{(i)})> t_{\bm X^{(i)}} \geq \max\big\{c_j(\bm x^{(i)})\big\}_{j=1\atop j\neq i}^N.
\end{align}

In deep learning, Convolutional Neural Networks (CNN) and Transformer have been applied to develop various popular classification models,
such as VGG \cite{simonyan2015very}, ResNet \cite{he2016deep}, ConvNeXt \cite{liu2022convnet}, ViT \cite{dosovitskiy2020image}, and DeiT \cite{touvron2021training}, etc.
These deep models usually take Full Connection (FC) or linear function as their classifiers.
A FC/linear classifier has a weight matrix $\bm W$ and a bias vector $\bm b$,
\begin{align}
\label{eq_fc_weight_bias}
\bm W & = \big(W_1, W_2,\cdots,W_N\big)\in \mathbb{R}^{M\times N},\\
\bm b & = \big(b_1, b_2,\cdots,b_N\big)^T \in \mathbb{R}^{N}.
\end{align}
For any sample feature $\bm x = \mathcal {M}(\bm X) \in \mathbb{R}^M$, the FC classifier $\mathcal C^{(\text{fc})}=\{c_i^{(\text{fc})}(W_i,b_i;\cdot)\}_{i=1}^N$ maps it to $N$ metrics,
\begin{align}
\label{eq_fc_values}
\big\{c_i^{(\text{fc})}(\bm x) = W_i^T\bm x + b_i\big\}_{i=1}^N.
\end{align}
These deep network models and their FC classifiers are trained on an annotated dataset (i.e., training dataset)
to tune their parameters for better feature representation.
In their training, they might not explicitly learn the threshold $t_{\bm X}$ to classify the sample $\bm X$,
while the threshold is doubtlessly available if the sample was correctly classified after the training.

\subsection{Sample-wise classification}
To achieve the highest possible classification accuracy in the testing or validation,
one should choose the adaptive thresholds that are tailored to the samples.


\begin{Def}
\label{def_sample_wise_classifition}
For a sample $\bm X^{(i)}\in \mathcal D$, it is \textbf{sample-wise classified} by the model $\mathcal M$ and $\mathcal C=\{c_i(\theta_i;\cdot)\}_{i=1}^N$,
if there exists a threshold $t_{\bm X^{(i)}}$ satisfying Eq. (\ref{eq_classification_condition_2}).

If all samples of $\mathcal D$ are sample-wise classified, the dataset $\mathcal D$ is said to be \textbf{sample-wise classified} by $\mathcal M$ and $\mathcal C$.

We define \textbf{sample-wise classification accuracy} as
\begin{align}
\label{eq_accuracy}
\mathcal {A}_{\text{SW}}({\mathcal M, \mathcal C}) = \frac{|\mathcal D({\mathcal M, \mathcal C})|}{|\mathcal D|} \times 100\%,
\end{align}
where $\mathcal D(\mathcal M, \mathcal C)$ consists of the samples sample-wise classified by $\mathcal M$ and $\mathcal C$ in $\mathcal D$,
i.e., $\mathcal D(\mathcal M, \mathcal C)$ is the biggest subset of $\mathcal D$ which is sample-wise classified by $\mathcal M$ and $\mathcal C$.
\hfill\RectangleBold
\end{Def}

Currently, the sample-wise classification and its accuracy $\mathcal {A}_{\text{SW}}$ are widely used in the evaluation of classification models and classifiers.
Obviously, this classification does not take into account the uniformity across sample features.

\subsection{Uniform classification}
In applications such as open-set classification and face recognition,
a unified threshold $t$, which is independent of the samples and separating the all samples' positive and negative classification metrics,
is more useful and efficient than the adaptive threshold.

\begin{Def}
\label{def_uniform_classifition}
For a dataset $\mathcal {D}$ with model $\mathcal M$ and classifier $\mathcal C=\{c_i(\theta_i;\cdot)\}_{i=1}^N$,
if there exists a unified threshold $t$ independent of samples in $\mathcal D$, satisfying Eq. (\ref{eq_classification_condition_2}) for the all samples in $\mathcal D$, i.e.,
\begin{align}
\label{eq_uniform_classification_condition}
\min\bigcup_{i=1}^N&\big\{c_i(\bm x^{(i)}):\bm x^{(i)}\in\mathcal F_i\big\} >t \nonumber\\
 &\geq \max\bigcup_{i=1}^N\bigcup_{j=1\atop j\neq i}^N\big\{c_j(\bm x^{(i)}):\bm x^{(i)}\in\mathcal F_i\big\},
\end{align}
$\mathcal D$ is defined as \textbf{uniformly classified} by $\mathcal M$ and $\mathcal C$ with $t$.

In contrary, with a fixed threshold $t$, $\mathcal M$ and $\mathcal C$ typically only classify a subset of $\mathcal D$,
we denote it as $\mathcal D(\mathcal M, \mathcal C; t)$, i.e., the biggest subset of $\mathcal D$ uniformly classified by $\mathcal M$ and $\mathcal C$ with $t$.
Then the ratio
\begin{align}
\label{eq_uniform_accuracy_0}
\mathcal A_{\text{Uni}} (\mathcal M, \mathcal C; t) = \frac{|\mathcal D(\mathcal M, \mathcal C; t)|}{|\mathcal D|} \times 100\%,
\end{align}
is the corresponding accuracy,
and the maximum ratio with varying thresholds, i.e.,
\begin{align}
\label{eq_uniform_accuracy}
\mathcal {A}_{\text{Uni}}(\mathcal M, \mathcal C) &= \max_{t\in\mathbb{R}} \mathcal A_{\text{Uni}} (\mathcal M, \mathcal C; t),
\end{align}
is defined as the \textbf{uniform classification accuracy}. 
\hfill\RectangleBold
\end{Def}

Usually, given dataset $\mathcal D$ with model $\mathcal M$ and classifier $\mathcal C$,
$\mathcal D(\mathcal M, \mathcal C;t')\neq \mathcal D(\mathcal M, \mathcal C;t'')$ for any two different threshold $t'$ and $t''$;
there exists at least one optimal threshold $t^*$, such that
\begin{align}
& t^* = \arg\max_{t\in\mathbb{R}}\big|\mathcal D(\mathcal M, \mathcal C;t)\big|,\\
& \mathcal {A}_{\text{Uni}}(\mathcal M, \mathcal C) = \mathcal {A}_{\text{Uni}}(\mathcal M, \mathcal C; t^*).
\end{align}

\subsection{Class-wise uniform classification}
Lying between the sample-wise classification and the uniform classification is class-wise uniform classification.

\begin{Def}
\label{def_class_wise_uniform_classifition}
Given a dataset $\mathcal {D}=\bigcup_{i=1}^N\mathcal D_i$ with model $\mathcal M$ and classifier $\mathcal C=\{c_i(\theta_i;\cdot)\}_{i=1}^N$,
if, for any category $i$, there exists a unified threshold $t_i$ uniformly classifying the $i$-th subset $\mathcal D_i$, i.e.,
\begin{align}
\label{eq_class_wise_uniform_classification_condition}
\min&\big\{c_i(\bm x^{(i)}):\bm x^{(i)}\in\mathcal F_i\big\} > t_i \nonumber\\
&\geq \max\bigcup_{j=1\atop j\neq i}^N\big\{c_j(\bm x^{(i)}):\bm x^{(i)}\in\mathcal F_i\big\},\quad \forall i,
\end{align}
then we say that $\mathcal {D}$ is \textbf{class-wise uniformly classified} by the model $\mathcal M$ and classifier $\mathcal C$
with thresholds $\{t_i\}_{i=1}^N$.

For $\forall i$, with a fixed threshold $t_i$, it is common that only a subset of $\mathcal D_i$ can be uniformly classified by $\mathcal M$ and $\mathcal C$,
and we denote it as $\mathcal D_i(\mathcal M, \mathcal C; t_i)$.
Then, the \textbf{class-wise uniform classification accuracy} of $\mathcal M$ and $\mathcal C$ on $\mathcal D$ is defined as
\begin{align}
\label{eq_class_wise_uniform_classification_accuracy}
\mathcal A_{\text{CW}} (\mathcal M, \mathcal C) &= \max_{\{t_i\}_{i=1}^N\in\mathbb{R}}\mathcal A_{\text{CW}}\big(\mathcal M, \mathcal C;\{t_i\}_{i=1}^N\big),
\end{align}
where
\begin{align}
\mathcal A_{\text{CW}}\big(\mathcal M, \mathcal C;\{t_i\}_{i=1}^N\big) &= \frac{\big|\bigcup_{i=1}^N \mathcal D_i(\mathcal M,\mathcal C; t_i)\big|}{|\mathcal D|}\\
            & = \frac{\sum_{i=1}^N\big|\mathcal D_i\big(\mathcal M, \mathcal C; t_i\big)\big|}{|\mathcal D|}
\end{align}
is the class-wise uniform classification accuracy of $\mathcal M$ and $\mathcal C$ on $\mathcal D$ with thresholds $\{t_i\}_{i=1}^N$.
\hfill\RectangleBold
\end{Def}

Similarly, there exists at least one set of optimal thresholds $\{t_i^*\}_{i=1}^N$, satisfying
\begin{align}
&\{t^*_i\}_{i=1}^N = \arg\max_{\{t_i\}_{i=1}^N\subset\mathbb{R}}\Big|\bigcup_{i=1}^N \mathcal D_i(\mathcal M,\mathcal C; t_i)\Big|,\\
&\mathcal A_{\text{CW}}\big(\mathcal M, \mathcal C\big) = \mathcal A_{\text{CW}}\big(\mathcal M, \mathcal C;\{t_i^*\}_{i=1}^N\big).
\end{align}

One can easily conclude that ``uniform classification'' $\Rightarrow$ ``class-wise uniform classification'' $\Rightarrow$ ``sample-wise classification'',
i.e., 
if $\mathcal M$ and $\mathcal C$ uniformly classify a dataset $\mathcal D$, they must class-wise uniformly classify $\mathcal D$,
and if $\mathcal M$ and $\mathcal C$ class-wise uniformly classify $\mathcal D$, they must sample-wise classify $\mathcal D$,
while both the oppositions are not necessarily the cases.

In general, given $\mathcal D$ with $\mathcal M$ and $\mathcal C$, we have
\begin{align}
|\mathcal D(\mathcal M,\mathcal C;t^*)|&= \Big|\bigcup_{i=1}^N \mathcal D_i(\mathcal M, \mathcal C; t^*)\Big| \\
            &\leq\Big|\bigcup_{i=1}^N \mathcal D_i(\mathcal M, \mathcal C; t_i^*)\Big| \\
            &\leq\Big|\bigcup_{i=1}^N \mathcal D_i(\mathcal M, \mathcal C)\Big| = |\mathcal D(\mathcal M,\mathcal C)|,
\end{align}
where $\mathcal D_i(\mathcal M, \mathcal C)$ is the subset of $\mathcal D_i$ containing its all samples which are sample-wise classified by $\mathcal M$ and $\mathcal C$.
Then,
\begin{align}
\label{eq_relation_3_accuracy}
\mathcal A_{\text{Uni}}&(\mathcal M, \mathcal C) \leq\mathcal A_{\text{CW}}(\mathcal M, \mathcal C) \leq \mathcal A_{\text{SW}}(\mathcal M, \mathcal C),
\end{align}
where $\mathcal {A}_{\text{SW}}(\mathcal M, \mathcal C)$ is the sample-wise classification accuracy defined in Eq. (\ref{eq_accuracy}).

\subsection{Uniform classification in the real-wold}
\subsubsection{Open-set classification}
For model $\mathcal M$ and classifier $\mathcal C$ trained on dataset $\mathcal D = \bigcup_{i=1}^N\mathcal D_i$ containing samples captured from $N$ categories,
in open-set classification, they would be applied to recognize samples captured from new category.
Suppose a sample $\bm X^{(N+1)}$ from category $N+1$, which is transformed to its feature $\bm x^{(N+1)} = \mathcal M(\bm X^{(N+1)})$.
Then, according to Eq. (\ref{eq_classification_condition_1}), the classifier $\mathcal C=\{c_i(\theta_i;\cdot)\}_{i=1}^N$ will predict its category ID as one in $\{1,2,\cdots,N\}$,
according to the largest classification metric $\max\{c_j(\bm x^{(N+1)})\}_{j=1}^N$,
which is undoubtedly wrong.

The maximum classification metric $\max\{c_j(\bm x)\}_{j=1}^N$ can be regarded as the response of model $\mathcal M$ and classifier $\mathcal C$ on the sample $\bm X$.
As $\mathcal M$ and $\mathcal C$ are trained on the samples from the $N$ categories,
the response on the sample $\bm X^{(N+1)}$ from the new category $N+1$ should be weak,
and that on the samples from the old categories would be stronger.
Therefore, in the open setting, to prevent incorrectly classifying samples from unknown categories into the closed set $\{1,2,\cdots,N\}$,
a fixed threshold $t$ should be chosen in advance,
and then the model and classifier could filter out these samples by comparing the responses with the threshold $t$.

It is notable that, for a sample from unknown category, $\bm X^{(N+1)}$,
its classification metric $\{c_i(\bm x^{(N+1)})\}_{i=1}^N$ generated by $\mathcal M$ and $\mathcal C$ are negative ones.
Therefore, the unified threshold $t$ is also expected to separate the positive classification metrics of the samples from the known categories $i\in\{1,2,\cdots,N\}$
and the negative ones of the samples for the above setting.
Clearly, the uniform classification is more suitable for the classification in the open setting than the sample-wise one.

\subsubsection{Face recognition}
With approximately eight billion people in the world, it is evidently challenging to gather a facial image dataset from every individual to train a face recognition model.
The commonly used facial dataset CASIA-WebFace comprises facial images of merely 10,575 individuals.
There remains an expectation that a face recognition model could accurately identify faces it hasn't encountered during its training,
which signifies that the face recognition constitutes a canonical open-set classification task.

In the practical application, the face recognition tasks are divided into $1\text{:}n$ identification and $1\text{:}1$ verification.
For the $1\text{:}n$ identification, it involves the identification of $n$ individuals, whose facial images may not be encompassed in the training dataset of the model $\mathcal M$.
One can first registers a standard image $\bm X_*^{(i)}$ for each individual $i$;
then, for an image $\bm X$ to be identified, calculates and compares the similarity of its feature $\bm x = \mathcal M (\bm X)$
with the $n$ standard features $\bm x_*^{(i)} = \mathcal M(\bm X_*^{(i)})$, respectively;
and, finally, classifies the image $\bm X$ into the category of the standard image that has the highest feature similarity with it.
If we denote the similarity of two features $\bm x_*^{(i)}, \bm x$ using a metric function $c(\bm x_*^{(i)};\bm x)$,
such as $c(\bm x_*^{(i)};\bm x)=\frac{\langle\bm x_*^{(i)},\bm x\rangle}{\|\bm x_*^{(i)}\|\|\bm x\|}$,
we will get a classifier $\mathcal C = \big\{c(\bm x_*^{(i)};\cdot)\big\}_{i=1}^n$, where the standard feature plays the role of function parameters.
Then, for an image $\bm X$ of a new individual, a unified threshold $t$
can be applied to effectively detect that this individual has not been registered,
as its $n$ (negative) metrics are smaller than the threshold, i.e.,
$ \max\{c(\bm x_*^{(i)};\bm x)\}_{i=1}^n < t$ with $\bm x = \mathcal M(\bm X)$.

For the $1\text{:}1$ verification, it verifies whether two face images are captured from the same individual or not.
In the practice, for any two images $\bm X_*, \bm X$, a unified threshold $\hat t$ is chosen in advance,
then the two images are considered to belong to the same individual,
if the similarity $c(\bm x_*; \bm x)$ is greater than the threshold;
otherwise, they are deemed to be face images of different individuals.

Both of face $1\text{:}n$ identification and $1\text{:}1$ verification
require the comparison of the unified threshold and the feature similarities of face images.
Therefore, to match better the application scenario of face recognition,
a unified threshold is expected to separate the all positive and negative classification metrics.




\section{Loss functions}\label{sec_loss_function}

Different classifiers requires distinct loss functions for the model training.
In this section, we will design the loss functions suitable to the different kind of classifications
and the supervised training of the model $\mathcal M$ and classifier $\mathcal C = \{c_i(\theta_i;\cdot)\}_{i=1}^N$.
In this section, we effectively rediscover the SoftMax loss and BCE loss.

\subsection{Sample-wise loss}
For any sample $\bm X^{(i)}\in \mathcal D$, to correctly classify it, it expects large positive classification metric $c_i(\bm x^{(i)})$ and small negative ones $c_j(\bm x^{(i)}), j\neq i$.
Therefore, 
a naive loss function could be reasonably designed as
\begin{align}
\label{eq_naive_loss}
\mathcal L_{\text{naive}}(\bm X^{(i)}) = -c_i(\bm x^{(i)}) + \frac{1}{N-1}\sum_{j=1\atop j\neq i}^N c_j(\bm x^{(i)}).
\end{align}
Using the famous inequality of arithmetic and geometric means\footnote{$\sqrt[n]{\prod_{i=1}^na_i} \leq \frac{1}{n}\sum_{i=1}^na_i$ for $a_i\geq0$.}, one can derive that
\begin{align}
\nonumber
\label{eq_soft_naive_loss_1}
\mathcal L_{\text{naive}}(\bm X^{(i)})
\leq&- \frac{N}{N-1}\log\frac{\exp\big({c_i(\bm x^{(i)})}\big)}{\sum_{j=1}^N\exp\big(c_j(\bm x^{(i)})\big)}\\
&- \frac{N\log N}{N-1}.
\end{align}
One can find the SoftMax loss in the right side of Eq. (\ref{eq_soft_naive_loss_1}),
\begin{align}
\label{eq_naive_softmax_loss}
\mathcal L_{\text{soft}}(\bm X^{(i)}) = -\log\frac{\exp\big(c_i(\bm x^{(i)})\big)}{\sum_{j=1}^N\exp\big(c_j(\bm x^{(i)})\big)},
\end{align}
then,
\begin{align}
\label{eq_soft_naive_loss_2}
\mathcal L_{\text{naive}}(\bm X^{(i)})
\leq& \frac{N}{N-1} \mathcal L_{\text{soft}}(\bm X^{(i)}) - \frac{N\log N}{N-1}.
\end{align}


For every sample $\bm X^{(i)}\in\mathcal D$,
the above two loss functions, $\mathcal L_{\text{naive}}$ and $\mathcal L_{\text{soft}}$,
only expect high positive classification metric and low negative ones,
which results an adaptive threshold to separate the sample.
In other words, these two loss functions only match the sample-wise classification.

\subsection{Uniform loss}
We here design the loss functions for the uniform classification,
and integrate the unified thresholds $t$ and $\{t_i\}_{i=1}^N$ defined in Eqs. (\ref{eq_uniform_classification_condition}) and (\ref{eq_class_wise_uniform_classification_condition}).
One will dramatically find that the loss functions are provided with BCE formulas, and the thresholds are evolved to the metric bias in these loss functions.

We start the loss design from the naive loss $\mathcal L_{\text{naive}}$,
 and first derive its three inequalities using the inequality of arithmetic and geometric means,
\begin{align}
\nonumber
\label{eq_loss_negative_term}
\mathcal L_{\text{naive}}(\bm X^{(i)})
\leq~&2\log\Big(1+\frac{\exp\big(\sum_{j=1\atop j\neq i}^N\frac{c_j(\bm x^{(i)})}{N-1}\big)}{\exp\big(c_i(\bm x^{(i)})\big)}\Big)\\
&- 2\log2,\\
\nonumber
\label{eq_loss_positive_term}
\mathcal L_{\text{naive}}(\bm X^{(i)})
\leq~&\frac{2}{N-1} \sum_{j=1\atop j\neq i}^N\log\Big(1+\frac{\exp\big(c_j(\bm x^{(i)})\big)}{\exp\big(c_i(\bm x^{(i)})\big)}\Big)\\
&-2\log 2,
\end{align}
and
\begin{align}
\nonumber
\label{eq_loss_negative_term_sum}
\sum_{i=1}^N \mathcal L_{\text{naive}}(\bm X^{(i)})
\leq& \frac{2}{N-1}\sum_{i=1}^N\sum_{j=1\atop j\neq i}^N\log\Big(1+\frac{\exp\big(c_j(\bm x^{(i)})\big)}{\exp\big(c_j(\bm x^{(j)})\big)}\Big)\\
&- 2N\log2.
\end{align}

To infer the loss functions for the uniform classification,
we here suppose that
the unified thresholds $t$ and $\{t_i\}_{i=1}^N$ exist and satisfy Eq. (\ref{eq_uniform_classification_condition}) and (\ref{eq_class_wise_uniform_classification_condition}).
Then, combining Eqs. (\ref{eq_loss_negative_term}), (\ref{eq_loss_positive_term}), and (\ref{eq_loss_negative_term_sum}),
one can get

\begin{align}
\nonumber
&\frac{N}{2}\mathcal L_{\text{naive}}(\bm X^{(i)})+N\log2\\
\leq &~ \log\Big(1+\frac{\e^{\sum_{j=1\atop j\neq i}^N\frac{c_j(\bm x^{(i)})}{N-1}}}{\e^{c_i(\bm x^{(i)})}}\Big)
\label{eq_uce_p1}
     + \sum_{j=1\atop j\neq i}^N\log\Big(1+\frac{\e^{c_j(\bm x^{(i)})}}{\e^{c_i(\bm x^{(i)})}}\Big)\\
\leq &~ \log\Big(1+\frac{\e^{\sum_{j=1\atop j\neq i}^N\frac{t}{N-1}}}{\e^{c_i(\bm x^{(i)})}}\Big)
\label{eq_uce_p2}
     + \sum_{j=1\atop j\neq i}^N\log\Big(1+\frac{\e^{c_j(\bm x^{(i)})}}{\e^{t}}\Big)\\
= &~ \log\Big(1+\e^{t-c_i(\bm x^{(i)})}\Big)
\label{eq_uce_p3}
     + \sum_{j=1\atop j\neq i}^N\log\Big(1+\e^{c_j(\bm x^{(i)})-t}\Big)
\end{align}
and
\begin{align}
\nonumber
&\frac{N}{2}\sum_{i=1}^N\mathcal L_{\text{naive}}(\bm X^{(i)})+N^2\log2\\
\label{eq_bce_p1}
\leq & \sum_{i=1}^N\Big[\log\Big(1+\frac{\e^{\sum_{j=1\atop j\neq i}^N\frac{c_j(\bm x^{(i)})}{N-1}}}{\e^{c_i(\bm x^{(i)})}}\Big)
        + \sum_{j=1\atop j\neq i}^N\log\Big(1+\frac{\e^{c_j(\bm x^{(i)})}}{\e^{c_j(\bm x^{(j)})}}\Big)\Big]\\
\label{eq_bce_p2}
\leq & \sum_{i=1}^N\Big[\log\Big(1+\frac{\e^{\sum_{j=1\atop j\neq i}^N\frac{t_i}{N-1}}}{\e^{c_i(\bm x^{(i)})}}\Big)
        + \sum_{j=1\atop j\neq i}^N\log\Big(1+\frac{\e^{c_j(\bm x^{(i)})}}{\e^{t_j}}\Big)\Big]\\
\label{eq_bce_p3}
= & \sum_{i=1}^N\Big[\log\big(1+\e^{t_i-c_i(\bm x^{(i)})}\big)
        + \sum_{j=1\atop j\neq i}^N\log\big(1+\e^{c_j(\bm x^{(i)})-t_j}\big)\Big].
\end{align}

The terms in the right sides of inequalities (\ref{eq_uce_p3}) and (\ref{eq_bce_p3}) both take the form of BCE,
but with different biases in their exponentials. Derived from these inequalities, we define \emph{\textbf{u}}nified threshold integrated BCE loss,
\begin{align}
\label{eq_bce_u_loss}
\nonumber
\mathcal L_{\text{bce-u}}(\bm X^{(i)}) = &\log\Big(1+\e^{b-c_i(\bm x^{(i)})}\Big)\\
     &+ \sum_{j=1\atop j\neq i}^N\log\Big(1+\e^{c_j(\bm x^{(i)})-b}\Big),
\end{align}
and \emph{\textbf{d}}iverse thresholds integrated BCE loss,
\begin{align}
\nonumber
\mathcal L_{\text{bce-d}}(\bm X^{(i)}) = &\log\Big(1+\e^{b_i-c_i(\bm x^{(i)})}\Big)\\
\label{eq_bce_d_loss}
&     + \sum_{j=1\atop j\neq i}^N\log\Big(1+\e^{c_j(\bm x^{(i)})-b_j}\Big),
\end{align}
where $b=t$ and $\{b_i=t_i\}_{i=1}^N$ are learnable parameters.

$\mathcal L_{\text{bce-u}}$ and $\mathcal L_{\text{bce-d}}$ share the similar expressions,
differing only in the bias term within the exponent.
Thought $\mathcal L_{\text{bce-d}}$ is designed for $N$-class classification tasks,
it can be easily derived from the binary cross-entropy (BCE) function used for binary classification tasks.
Next, we will discuss whether these two losses are suitable for the uniform classification and class-wise uniform classification.

\subsection{Convergence of the threshold}
In the design of losses $\mathcal L_{\text{bce-u}}$ and $\mathcal L_{\text{bce-d}}$,
we suppose that the unified thresholds $t$ and $\{t_i\}_{i=1}^N$ exist.
These thresholds are evolved into the learnable biases $b$ and $\{b_i\}_{i=1}^N$ in the loss functions,
which raises a question: will the biases converge to the thresholds after the training of model $\mathcal M$ and classifier $\mathcal C$ using the losses?

\begin{Cor}
For the dataset $\mathcal D=\bigcup_{i=1}^N\mathcal D_i$
with the model $\mathcal M$ and classifier $\mathcal C=\{c_i(\theta_i;\cdot)\}$,
we suppose that $c_i(\theta_i;\cdot)$ has lower bound $A$ and upper bound $B$ ($B>A$) for $\forall~i$, and
\begin{align}
    \label{eq_bce_u_condition}
    2\leq N<\frac{\e^{B-A}+3}{2}.
\end{align}
If the model and classifier are perfectly trained on $\mathcal D$ using $\mathcal L_{\text{bce-u}}(\bm X^{(i)})$, i.e.,
\begin{enumerate}[~~(i).]
  \item for $\forall \bm X^{(i)}\in \mathcal D$, its positive metrics $c_i(\bm x^{(i)})$ tends to $B$ and the negative ones $c_j(\bm x^{(i)})$ tends to $A$,
  \item the loss $\mathcal L_{\text{bce-u}}(\bm X^{(i)})$ has reached its minimum point,
\end{enumerate}
then the final learned bias $b$ in $\mathcal L_{\text{bce-u}}$ is a unified threshold satisfying Eq. (\ref{eq_uniform_classification_condition}).

\emph{\textbf{Proof}:}
As the loss function $\mathcal L_{\text{bce-u}}$ is differentiable in terms of bias $b$,
its minimum point must be the bias' stationary point, which satisfies
\begin{align}
\label{eq_bias_stationary_point}
0 &=\frac{\partial \mathcal L_{\text{bce-u}}}{\partial b} =\frac{\e^{-c_i(\bm x^{(i)})+b}}{1+\e^{-c_i(\bm x^{(i)})+b}}
- \sum_{j=1\atop j\neq i}^N\frac{\e^{c_j(\bm x^{(i)})-b}}{1+\e^{c_j(\bm x^{(i)})-b}}\\
&\overset{\text{(i)}}{=} \frac{\e^{-B +b}}{1+\e^{-B +b}} - \sum_{j=1\atop j\neq i}^N\frac{\e^{A-b}}{1+\e^{A-b}}\\
\label{eq_unified_b}
\Rightarrow b&=\log\frac{(N-2)\e^A + \sqrt{(N-2)^2\e^{2A}+4(N-1)\e^{A+B}}}{2}.
\end{align}
For the $b$ in Eq. (\ref{eq_unified_b}),
\begin{align}
\label{eq_bias_stationary_point_l}
b& > \log\frac{\sqrt{4\e^{A+B}}}{2} \geq \log\frac{\sqrt{4\e^{2A}}}{2} = A,~\text{if}~N\geq2;\\
\nonumber
b& < \log\frac{\frac{\e^{B-A}-1}{2}\e^A + \sqrt{(\frac{\e^{B-A}-1}{2})^2\e^{2A}+4(\frac{\e^{B-A}+1}{2})\e^{A+B}}}{2}\\
\label{eq_bias_stationary_point_r}
 & = \log\frac{\frac{\e^B-\e^A}{2} + \sqrt{(\frac{3\e^B}{2}+\frac{\e^A}{2})^2}}{2}=B,~\text{if}~N<\frac{\e^{B-A}+3}{2}.
\end{align}

Combining Eqs. (\ref{eq_bias_stationary_point_l}) and (\ref{eq_bias_stationary_point_r}),
one can conclude that the learned bias $b$ have separated the all samples' positive classification metrics and negative ones,
i.e., the learned bias is a unified threshold, satisfying Eq. (\ref{eq_uniform_classification_condition}).
\hfill\RectangleBold
\end{Cor}

\begin{table*}[htbp]
  \centering
  \caption{Twelve loss functions}\label{Tab_12_losses}
  \begin{tabular}{c|ccc c r}\hline
  Loss                      & Formula   & Classifier    & Bias                      & \multicolumn{2}{c}{Loss function expression}\\\hline\hline
  $L_{\text{soft-0}}$       & SoftMax   & $W_i^T\bm x$  & 0
            & $- \log\big[\exp\big(W_i^T\bm x^{(i)}\big)\big/\sum_{j=1}^N\exp\big(W_j^T\bm x^{(i)}\big)\big]$
            &\newcounter{losssoftz} \stepcounter{equation} \setcounter{losssoftz}{\value{equation}} \hypertarget{losssoftz}{(\thelosssoftz)} \\\hline
  $ L_{\text{soft-d}}$       & SoftMax   & $W_i^T\bm x$  & $b_i$                     &
            $ -\log\big[\exp\big(W_i^T\bm x^{(i)}+b_i\big)\big/\sum_{j=1}^N\exp\big(W_j^T\bm x^{(i)}+b_j\big)\big]$
            &\newcounter{losssoftd} \stepcounter{equation} \setcounter{losssoftd}{\value{equation}} \hypertarget{losssoftd}{(\thelosssoftd)}\\\hline
  $ L_{\text{soft-u}}$       & SoftMax   & $W_i^T\bm x$  & $b$
            & $-\log\big[\exp\big(W_i^T\bm x^{(i)}+b\big)\big/\sum_{j=1}^N\exp\big(W_j^T\bm x^{(i)}+b\big)\big]$
            &\newcounter{losssoftu} \stepcounter{equation} \setcounter{losssoftu}{\value{equation}} \hypertarget{losssoftu}{(\thelosssoftu)}\\\hline
  $ L_{\text{soft-n0}}$      & SoftMax   & $\gamma\cos(W_i, \bm x)=\frac{\gamma W_i^T\bm x}{\|W_i\|\|\bm x\|}$     & 0
            & $-\log\big[\e^{\gamma \cos(W_i,\bm x^{(i)})}\big/\sum_{j=1}^N\e^{\gamma \cos(W_j,\bm x^{(i)})}\big]$
            &\newcounter{losssoftnz} \stepcounter{equation} \setcounter{losssoftnz}{\value{equation}} \hypertarget{losssoftnz}{(\thelosssoftnz)}\\\hline
  $ L_{\text{soft-nd}}$      & SoftMax   & $\gamma\cos(W_i, \bm x)=\frac{\gamma W_i^T\bm x}{\|W_i\|\|\bm x\|}$     & $b_i$
            & $-\log\big[\e^{\gamma \cos(W_i,\bm x^{(i)})-b_i}\big/\sum_{j=1}^N\e^{\gamma \cos(W_j,\bm x^{(i)})-b_j}\big]$
            &\newcounter{losssoftnd} \stepcounter{equation} \setcounter{losssoftnd}{\value{equation}} \hypertarget{losssoftnd}{(\thelosssoftnd)}\\\hline
  $ L_{\text{soft-nu}}$      & SoftMax   & $\gamma\cos(W_i, \bm x)=\frac{\gamma W_i^T\bm x}{\|W_i\|\|\bm x\|}$     & $b$
            & $-\log\big[\e^{\gamma \cos(W_i,\bm x^{(i)})-b}\big/\sum_{j=1}^N\e^{\gamma \cos(W_j,\bm x^{(i)})-b}\big]$
            &\newcounter{losssoftnu} \stepcounter{equation} \setcounter{losssoftnu}{\value{equation}} \hypertarget{losssoftnu}{(\thelosssoftnu)}\\\hline
  $ L_{\text{bce-0}}$       & BCE   & $W_i^T\bm x$  & 0
            & $\log\Big(1+\e^{-W_i^T\bm x^{(i)}}\Big) + \sum_{j=1\atop j\neq i}^N\log\Big(1+\e^{W_j^T\bm x^{(i)}}\Big)$
            &\newcounter{lossbcez} \stepcounter{equation} \setcounter{lossbcez}{\value{equation}} \hypertarget{lossbcez}{(\thelossbcez)}\\\hline
  $ L_{\text{bce-d}}$       & BCE   & $W_i^T\bm x$  & $b_i$
            & $\log\Big(1+\e^{-W_i^T\bm x^{(i)}-b_i}\Big) + \sum_{j=1\atop j\neq i}^N\log\Big(1+\e^{W_j^T\bm x^{(i)}+b_j}\Big)$
            &\newcounter{lossbced} \stepcounter{equation} \setcounter{lossbced}{\value{equation}} \hypertarget{lossbced}{(\thelossbced)}\\\hline
  $ L_{\text{bce-u}}$       & BCE   & $W_i^T\bm x$  & $b$
            & $\log\Big(1+\e^{-W_i^T\bm x^{(i)}-b}\Big) + \sum_{j=1\atop j\neq i}^N\log\Big(1+\e^{W_j^T\bm x^{(i)}+b}\Big)$
            &\newcounter{lossbceu} \stepcounter{equation} \setcounter{lossbceu}{\value{equation}} \hypertarget{lossbceu}{(\thelossbceu)}\\\hline
  $ L_{\text{bce-n0}}$       & BCE   & $\gamma\cos(W_i, \bm x)=\frac{\gamma W_i^T\bm x}{\|W_i\|\|\bm x\|}$  & 0
            & $\log\Big(1+\e^{-\gamma \cos(W_i,\bm x^{(i)})}\Big) + \sum_{j=1\atop j\neq i}^N\log\Big(1+\e^{\gamma \cos(W_j,\bm x^{(i)})}\Big)$
            &\newcounter{lossbcenz} \stepcounter{equation} \setcounter{lossbcenz}{\value{equation}} \hypertarget{lossbcenz}{(\thelossbcenz)}\\\hline
  $ L_{\text{bce-nd}}$       & BCE   & $\gamma\cos(W_i, \bm x)=\frac{\gamma W_i^T\bm x}{\|W_i\|\|\bm x\|}$  & $b_i$
            & $\log\Big(1+\e^{-\gamma \cos(W_i,\bm x^{(i)})+b_i}\Big) + \sum_{j=1\atop j\neq i}^N\log\Big(1+\e^{\gamma \cos(W_j,\bm x^{(i)})-b_j}\Big)$
            &\newcounter{lossbcend} \stepcounter{equation} \setcounter{lossbcend}{\value{equation}} \hypertarget{lossbcend}{(\thelossbcend)}\\\hline
  $ L_{\text{bce-nu}}$       & BCE   & $\gamma\cos(W_i, \bm x)=\frac{\gamma W_i^T\bm x}{\|W_i\|\|\bm x\|}$  & $b$
            & $\log\Big(1+\e^{-\gamma \cos(W_i,\bm x^{(i)})+b}\Big) + \sum_{j=1\atop j\neq i}^N\log\Big(1+\e^{\gamma \cos(W_j,\bm x^{(i)})-b}\Big)$
            &\newcounter{lossbcenu} \stepcounter{equation} \setcounter{lossbcenu}{\value{equation}} \hypertarget{lossbcenu}{(\thelossbcenu)}\\\hline
  \end{tabular}
\end{table*}
The above corollary indicates that $\mathcal L_{\text{bce-u}}$ is the loss suitable for the uniform classification.
However, unlike $\mathcal L_{\text{bce-u}}$, the $\mathcal L_{\text{bce-d}}$ defined by Eq. (\ref{eq_bce_d_loss}) cannot learn the class-wise unified threshold $t_i$,
though the thresholds are adopted in its design.
In fact, during the model optimization using $\mathcal L_{\text{bce-d}}$,
it causes the positive metric $c_i(\bm x^{(i)})$ to be greater than the bias $b_i$ and the negative metric $c_j(\bm x^{(i)})$ to be less than $b_j$,
for any sample feature $\bm x^{(i)}\in \mathcal F_i$,
resulting in thresholds $\{b_i = \hat{t}_i\}_{i=1}^N$ satisfying
\begin{align}
\label{eq_class_wise_uniform_classification_condition_2}
\min&\big\{c_i(\bm x^{(i)}):\bm x^{(i)}\in\mathcal F_i\big\} > \hat t_i \nonumber\\
&\geq \max\bigcup_{j=1\atop j\neq i}^N\big\{c_i(\bm x^{(j)}):\bm x^{(j)}\in\mathcal F_j\big\},\quad \forall i.
\end{align}
For the $i$-th class, we refer to $c_j(\bm x^{(i)})$ as \textbf{type I negative metric}, which measures the distance or similarity of sample features from the category $i$ to other categories,
while \(c_i(\bm x^{(j)})\) is termed as \textbf{type II negative metric}, assessing that of sample features from other categories to the class $i$.

The class-wise uniform classification defined by Eq. (\ref{eq_class_wise_uniform_classification_condition}) requires the thresholds
that uniformly distinguishes all positive metrics and all Type I negative metrics of all samples in each category.
However, $\mathcal L_{\text{bce-d}}$ learns the thresholds that uniformly distinguishes all positive metrics and all Type II negative metrics.
In other words, $\mathcal L_{\text{bce-d}}$ is not the loss suitable for the class-wise uniform classification.

\begin{table*}[t]
  \centering
  \caption{The classification performances of twelve loss functions on ImageNet-1K}\label{Tab_imagenet1k_1}
  \setlength{\tabcolsep}{1.00mm}{
  \begin{tabular}{c|c||c| ccc| ccc| ccc| ccc}\hline
  \multirow{3}{*}{Model} & \multirow{3}{*}{Accuracy} &pre-training&\multicolumn{12}{c}{fine-tuning}\\\cline{3-15}
  &        &$L_{\text{soft-d}}$    &$L_{\text{soft-0}}$    &\makecell[c]{$L_{\text{soft-d}}$\\(baseline)}    &$L_{\text{soft-u}}$    &$L_{\text{soft-n0}}$    &$L_{\text{soft-nd}}$    &$L_{\text{soft-nu}}$
                                    &\color{black}$L_{\text{bce-0}}$    &\color{black}$L_{\text{bce-d}}$    &\color{black}$L_{\text{bce-u}}$    &$L_{\text{bce-n0}}$    &$L_{\text{bce-nd}}$    &$L_{\text{bce-nu}}$\\\hline\hline
  \multirow{4}{*}{ResNet50}&$\mathcal A_{\text{SW}}$  &75.80  &76.82    &76.74    &76.87    &77.21    &77.07    &77.22    &77.00    &77.12    &76.99    &77.34    &77.38    &\textbf{77.50}\\
&$\mathcal A_{\text{CW}}$  &46.32 &45.39    &45.51    &45.55    &49.46    &49.46    &49.53    &69.65    &69.56    &69.69    &70.59    &70.47    &\textbf{70.98}\\
&$\mathcal A_{\text{Uni}}$ &38.47 &34.38    &34.48    &34.44    &39.56    &39.69    &39.53    &67.04    &66.92    &67.16    &67.97    &67.93    &\textbf{68.48}\\
&$t^*$      &15.084 &12.490   &12.125   &9.545   &16.650   &16.401   &4.065   &$-$0.393   &$-$0.403   &$-$0.398   &$-$0.242   &$-$0.234   &$-$0.196\\\hline

  \multirow{4}{*}{ResNet101}&$\mathcal A_{\text{SW}}$  &77.26   &78.39    &78.47    &78.42    &78.62    &78.72    &78.59    &78.92    &78.88    &79.01    &79.11    &79.06    &\textbf{79.16}\\
&$\mathcal A_{\text{CW}}$  &50.14  &49.43    &49.23    &49.40    &51.72    &51.93    &51.84    &72.79    &72.87    &72.94    &73.37    &73.42    &\textbf{73.75} \\
&$\mathcal A_{\text{Uni}}$ &42.64 &39.17    &38.85    &38.77    &42.60    &42.81    &42.37    &70.43    &70.46    &70.61    &71.12    &71.12    &\textbf{71.41}\\
&$t^*$      &14.987 &12.266   &12.307   &9.136   &16.431   &16.633   &4.060   &$-$0.451   &$-$0.387   &$-$0.322   &$-$0.030   &$-$0.137   &$-$0.153\\\hline

  \multirow{4}{*}{DenseNet161}&$\mathcal A_{\text{SW}}$  &77.38&78.51    &78.58    &78.64    &78.05    &78.05    &77.94    &79.07    &79.19    &79.15    &79.24    &79.10    &\textbf{79.29}\\
&$\mathcal A_{\text{CW}}$  &49.92 &51.40    &51.28    &51.23    &51.24    &51.23    &51.18    &71.63    &71.67    &71.57    &71.47    &71.45    &\textbf{71.91}\\
&$\mathcal A_{\text{Uni}}$ &42.74 &43.62    &43.56    &43.67    &42.36    &42.27    &42.65    &68.96    &69.08    &69.04    &68.61    &68.91    &\textbf{69.15}\\
&$t^*$      &15.035 &11.867   &12.011   &8.398   &15.709   &15.799   &3.118   &$-$0.290   &$-$0.273   &$-$0.360   &$-$0.419   &$-$0.376   &$-$0.314\\\hline

  \end{tabular}}
\end{table*}
\subsection{Twelve loss functions}
As the aforementioned analysis, besides the function formula,
the classifier $\mathcal C = \{c_i(\theta_i;\cdot)\}_{i=1}^N$ and the bias mode are key factors of the loss functions.
In this paper, using SoftMax and BCE formulas, we compare twelve losses
with two kinds of linear classifier,
\begin{enumerate}[~~~i.]
  \item the ordinary linear functions:
  \begin{align}
        \mathcal C^{(\text{lin})} = \big\{c_i^{(\text{lin})}(W_i;\bm x) = W_i^T\bm x\big\}_{i=1}^N,~~~
    \end{align}
  \item the normalized linear functions with a scaling factor:
  \begin{align}
        \mathcal C^{(\text{nor})} &= \Big\{c_i^{(\text{nor})}(W_i;\bm x) = \frac{\gamma W_i^T\bm x}{\|W_i\|\|\bm x\|}\Big\}_{i=1}^N,
  \end{align}
\end{enumerate}
and three bias modes,
\begin{enumerate}[~~~i.]
  \item zero bias, i.e., setting $b_i = 0$ for $i=1,2,\cdots,N$,
  \item diverse bias, i.e., using diverse parameters $b_i$,
  \item unified bias, i.e., using a unified parameter $b$.
\end{enumerate}
The metric functions of classifier $\mathcal C^{(\text{lin})}$ are unbounded, while the range of the metric functions of $\mathcal C^{(\text{nor})}$ is $[-\gamma,\gamma]$,
and $\gamma$ is a scaling factor expanding the function value range.

Table \ref{Tab_12_losses} lists the expressions of the twelve losses.
$L_{\text{soft-d}}$ is the commonly used SoftMax/cross-entropy loss in the general classification tasks;
$L_{\text{soft-0}}$ and $ L_{\text{soft-u}}$ are its two similars with zero bias and unified bias,
which have not received enough attention in the previous studies.
$ L_{\text{soft-n0}}, L_{\text{soft-nd}}, L_{\text{soft-nu}}$ are three normalized SoftMax losses with different bias modes,
which have been evolved into various high-performance losses in the field of face recognition.
In theory, $L_{\text{soft-0}} = L_{\text{soft-u}}$ and $L_{\text{soft-n0}} = L_{\text{soft-nu}}$ for $\forall~b$.
In experiments, they perform very similarly on large-scale datasets like ImageNet-1K.
However, their performance exhibits more noticeable differences on smaller-scale datasets such as CUB \cite{wah2011caltech}.
For these six SoftMax-based losses, there is no evidence indicating any connection between their biases and the classification threshold.
Thus, we consider them all to belong to the sample-wise loss.

$ L_{\text{bce-0}}, L_{\text{bce-d}}, L_{\text{bce-u}}$ are the ordinary BCE losses with different bias modes.
When using them to train a classification model, it actually treats an $N$-class classification task as $N$ individual binary classification tasks.
Their classifiers $\big\{W_i^T\bm x\big\}_{i=1}^N$ are unbounded, ranging from the negative infinity to the positive infinity.
According to the previous analysis, there is a close connection between their biases and the classification thresholds.

$L_{\text{bce-n0}}, L_{\text{bce-nd}}, L_{\text{bce-nu}}$ are three normalized BCE losses.
According to the previous analysis, when $\gamma$ is large enough, their metric bias, i.e., $b_i$ and $b$, can learn the classification thresholds.
$L_{\text{bce-n0}}$ employs a fixed bias $b=0$, equivalent to integrating a fixed unified threshold $t=0$;
whereas $L_{\text{bce-nd}}$ integrates a learnable unified threshold $t_i = \frac{b_i}{\gamma}$ separately for each class $i$.
$L_{\text{bce-nu}}$ is our UCE loss proposed in \cite{zhou2023uniface}, which is an instance of the $\mathcal L_{\text{bce-u}}$ in Eq. (\ref{eq_bce_u_loss}),
integrating a learnable unified threshold $t=\frac{b}{\gamma}$ for the whole dataset.
In practice, it is crucial to choose an appropriate value for the scalar $\gamma$ in the training with these three losses.
When $\gamma$ is too large, the loss will easily diverge towards infinity, resulting in overflow and optimization failure.
In contrary, if $\gamma$ is too small, the normalized BCE loss function has a relatively large minimum,
and it is prone to reaching this minimum before the model has been sufficiently optimized, resulting in poor performance,
which is consistent with the inequality (\ref{eq_bce_u_condition}).

\section{Experiments}
We implement the aforementioned twelve losses
during the training of models for six classification tasks, to evaluate their performance in the three different kinds of classification.
The sample-wise and uniform classification accuracies will be employed in the evaluation.

\subsection{Image classification on ImageNet}

Using the twelve loss functions in Table \ref{Tab_12_losses},
we first train three widely used deep networks, including ResNet50, ResNet101 \cite{he2016deep}, and DenseNet161 \cite{huang2017densely}, on ImageNet-1K \cite{krizhevsky2012imagenet}.
The training set of ImageNet-1K comprises 1.2M images from 1,000 distinct classes, while its validation set consists of 50K images, with 50 images per category.
ImageNet-1K has been widely employed in assessing the performance of diverse deep classification models.

On ImageNet-1K, we do not from scratch train the three deep networks.
Instead, we employ their pre-trained models, which have undergone training for 90 epochs,
and perform fine-tuning using the twelve loss functions, respectively.
In each fine-tuning, we set the initial learning rate to $10^{-4}$ with a cosine decay strategy,
and retrain the model for 30 epochs on the training set.
We empirically set $\gamma=96$ for the normalized SoftMax and BCE losses.
Table \ref{Tab_imagenet1k_1} presents their classification accuracies on the validation set and the optimal uniform classification threshold $t^*$.
In computing the three kinds of accuracy and searching the optimal threshold $t^*$,
we adopt the classification metrics with the learned biases.

We take the results of $L_{\text{soft-d}}$ as the baseline,
due to the fact that it is the most commonly used loss function in the classification.
One can find that from Table \ref{Tab_imagenet1k_1}, for the \textbf{three ordinary SoftMax losses}, i.e., $L_{\text{soft-0}}$, $L_{\text{soft-d}}$, and $L_{\text{soft-u}}$,
they perform very similarly on ImageNet-1K, with only a difference of approximately 0.1\% for each accuracy of the three deep models in terms of $\mathcal A_{\text{SW}}$.
Compared with the baseline, \textbf{the normalized SoftMax losses}, i.e., $L_{\text{soft-n0}}$, $L_{\text{soft-nd}}$, and $L_{\text{soft-nu}}$,
improve the three accuracies of ResNet50 and ResNet101, but reduce that of DenseNet161.
The inconsistent performance gains indicate that neither the ordinary SoftMax nor the normalized one is the best choice for the classification.
Furthermore, it can be observed that the uniform classification accuracies $\mathcal A_{\text{Uni}}$ and $\mathcal A_{\text{CW}}$
of the SoftMax and normalized SoftMax loss do not exceed 40\% and 50\%, respectively,
which are significantly lower than their sample-wise classification accuracies $\mathcal A_{\text{SW}}$,
indicating that they are indeed not suitable losses for the uniform classification.

For \textbf{the ordinary BCE losses}, $L_{\text{bce-0}}$, $L_{\text{bce-d}}$, and $L_{\text{bce-u}}$,
their sample-wise accuracy $\mathcal A_{\text{SW}}$ is always superior to that of the three ordinary SoftMax losses,
but not consistently superior to that of the normalized SoftMax.
For instance, $\mathcal A_{\text{SW}}$ of ResNet50 trained using $L_{\text{bce-d}}$ reaches 77.12\%, slightly lower than that of $L_{\text{soft-nu}}$ (77.22\%).
However, the two uniform accuracies have been significantly improved by over 20\%.
For example, $\mathcal A_{\text{CW}}$ and $\mathcal A_{\text{Uni}}$ of ResNet50 trained with $L_{\text{bce-d}}$ is 69.56\% and 66.92\%,
showing improvements of 24.05\% and 32.44\%, respectively, compared to $L_{\text{soft-d}}$.
When using \textbf{the normalized BCE losses}, the three accuracies for the all three models are further improved, with $L_{\text{bce-nu}}$ consistently achieving the optimal results.

For \textbf{the three types of accuracy} across different models, some facts are worth noting.
Firstly, for each model trained with different loss function, the three accuracies,
i.e., $\mathcal A_{\text{SW}}$, $\mathcal A_{\text{CW}}$, and $\mathcal A_{\text{Uni}}$,
satisfy the inequality (\ref{eq_relation_3_accuracy}), aligning with our analysis.
Secondly, despite the stable improvement in the model performance with the three accuracies by the BCE losses, their enhancements are varying.
The gains using normalized BCE losses on these metrics decrease from $\mathcal A_{\text{Uni}}$, $\mathcal A_{\text{CW}}$, to $\mathcal A_{\text{SW}}$.
Taking ResNet50 for example, $L_{\text{bce-nu}}$ raises its $\mathcal A_{\text{Uni}}$ and $\mathcal A_{\text{CW}}$ from 34.48\% and 45.51\% to 68.48\% and 70.98\%,
resulting in improvements of 34.00\% and 25.47\%, respectively.
However, it increases $\mathcal A_{\text{SW}}$ from 76.74\% to 77.50\%, with a modest improvement of 0.76\%.
Thirdly, while the uniform classification is stronger than the sample-wise classification,
it does not imply that a model with higher uniform accuracy will necessarily have higher sample-wise accuracy.
For instance, the ResNet50 trained with the three ordinary BCE achieves significantly higher uniform accuracy
than that trained with the SoftMax losses, but the former's sample-wise accuracy is slightly lower than that of ResNet50 trained with the normalized SoftMax.
These results stem from the fact that the uniform classification and the sample-wise classification are inherently different,
while the former requires the global separability of the dataset, and the latter demands local, sample-wise separability.

\begin{table}[!tbph]
  \centering
  \caption{The classification results of ResNet50 trained by $L_{\text{bce-nu}}$ with various $\gamma$ on ImageNet-1K.}\label{Tab_gamma_resnet50-bce-nu}
  \setlength{\tabcolsep}{0.80mm}{
  \begin{tabular}{c|c ccc cccc c}\hline

$\gamma$                    &1        &2       &3        &4        &5      &6      &7      &8      &12\\\hline
$\mathcal A_{\text{SW}}$    &0.10    &0.10    &8.07    &24.30    &59.86    &72.83    &74.26    &74.60  &76.07\\
$\mathcal A_{\text{CW}}$    &0.10    &0.10    &3.34    &13.04    &25.85    &68.60    &70.47    &70.89  &72.72\\
$\mathcal A_{\text{Uni}}$   &0.10    &0.10    &0.72    &4.08    &16.51    &62.41    &65.40    &66.37   &70.21\\
$t^*$                       &$-$1.89   &$-$6.46   &$-$5.13   &$-$4.11   &$-$5.38   &$-$1.11   &$-$1.08   &$-$1.02   &$-$0.49\\\hline\hline
$\gamma$                    &16     &24     &32     &48     &64     &80     &96         &112    &128 \\\hline
$\mathcal A_{\text{SW}}$    &76.60    &77.01    &77.27    &77.46    &\textbf{77.61}    &77.50    &77.50     &77.48    &77.44\\
$\mathcal A_{\text{CW}}$    &\textbf{72.94}    &72.58    &72.26    &71.86    &71.49    &71.14    &70.98    &70.47    &70.32\\
$\mathcal A_{\text{Uni}}$   &\textbf{70.64}    &70.19    &69.84    &69.36    &69.08    &68.69    &68.48    &68.15    &68.14 \\
$t^*$                   &$-$0.46   &$-$0.36   &$-$0.38   &$-$0.46   &$-$0.48   &$-$0.26   &$-$0.20    &$-$0.38   &$-$0.37\\\hline\hline
$\gamma$                    &144    &160    &176    &192    &208    &256    &512\\\hline
$\mathcal A_{\text{SW}}$        &77.41    &77.33    &77.27    &77.39    &0.10    &73.52    &73.52\\
$\mathcal A_{\text{CW}}$  &70.76    &70.67      &70.25    &70.27    &0.00    &49.58    &49.58\\
$\mathcal A_{\text{Uni}}$ &67.95    &67.80    &67.71    &67.73    &0.00    &39.20    &39.20\\
$t^*$                   &$-$0.29   &$-$0.25   &$-$0.29   &$-$0.29   &-   &75.87   &160.95\\\hline\hline
  \end{tabular}}
\end{table}
\begin{figure}[!tbhp]
	\centering
    \includegraphics*[scale=0.388, viewport=50 2 692 288]{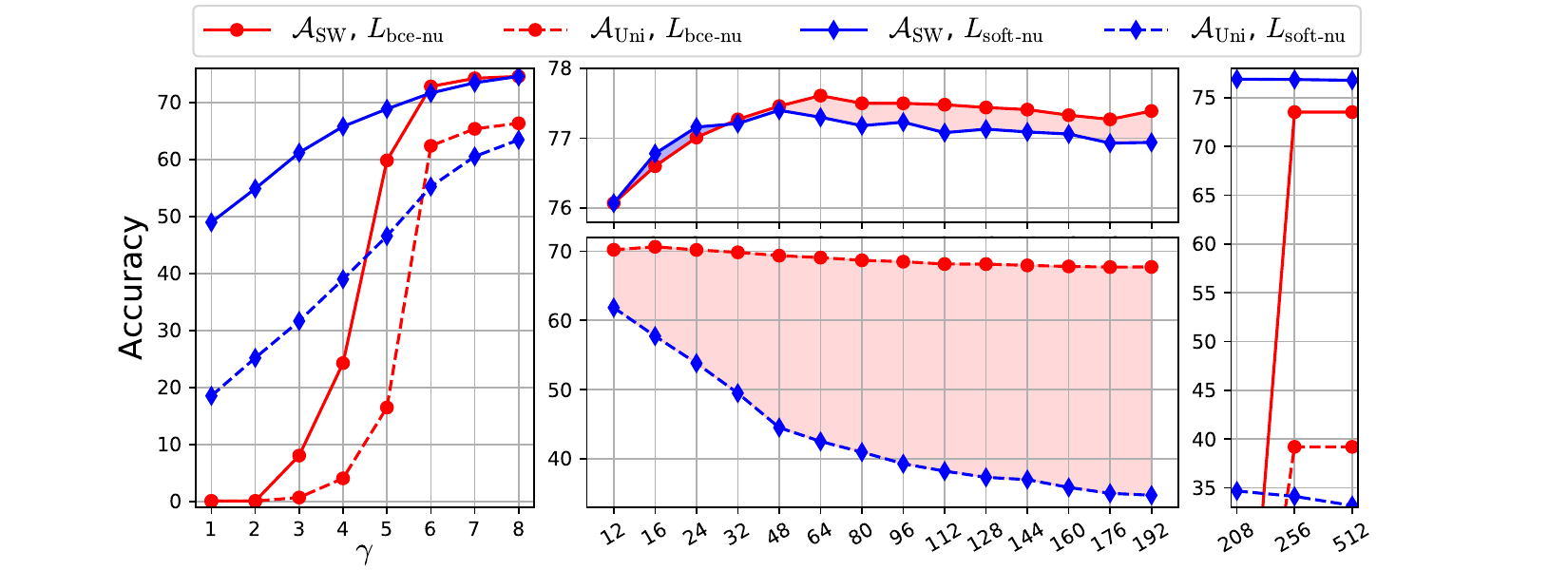}
	\caption{The visual comparison of performance of ResNet50 trained by $L_{\text{soft-nu}}$ and $L_{\text{bce-nu}}$ with various $\gamma$ on ImageNet-1K.
    Although $L_{\text{bce-nu}}$ performs poorly when $\gamma$ is too small or too large, for $\gamma$ varying in $[32,192]$,
    its uniform accuracy is much higher than $L_{\text{soft-nu}}$, and its sample-wise accuracy is slightly higher as well.}
	\label{fig_resnet50_accuracy_gamma}
\end{figure}
\textbf{The scalar factor $\gamma$} is a crucial parameter, when applying the normalized losses to train the models.
We select 25 different values for $\gamma$ and train ResNet50 using $L_{\text{soft-nu}}$ and $L_{\text{bce-nu}}$, respectively.
Table \ref{Tab_gamma_resnet50-bce-nu} presents the accuracies after training with $L_{\text{bce-nu}}$.
Fig. \ref{fig_resnet50_accuracy_gamma} visually compares the sample-wise accuracy $\mathcal A_{\text{SW}}$ and uniform accuracy $\mathcal A_{\text{Uni}}$ of Resnet50
trained with $L_{\text{soft-nu}}$ and $L_{\text{bce-nu}}$ as $\gamma$ varies.
From both Table \ref{Tab_gamma_resnet50-bce-nu} and Fig. \ref{fig_resnet50_accuracy_gamma},
one can find that ResNet50 trained with $L_{\text{bce-nu}}$ is unable to classify the ImageNet-1K validation set, when $\gamma$ is set to 1 or 2,
resulting in an accuracy of $1/N = 1/1000 = 0.1\%$.
Now, the inequality (\ref{eq_bce_u_condition}) does not hold.
In fact, the minimum value of BCE loss is relatively large when $\gamma$ is very small.
For example, the minimum value of $L_{\text{bce-nu}}$ is 5.91 with $\gamma =1$;
inspecting the training logs, we find that the loss value on the training set decreases from 693.17 to 12.14,
while the loss on the validation set always remains at 6.91.

As $\gamma$ increases from 3 to 6, the performance of model trained by $L_{\text{bce-nu}}$ rapidly improves in exponential,
with $\mathcal A_{\text{SW}}$ and $\mathcal A_{\text{Uni}}$ reaching 72.83\% and 62.41\%, respectively.
Subsequently, with further increases in $\gamma$, these metrics slowly improve to 77.61\% ($\gamma=64$) and 70.64\% ($\gamma=12$).
However, when $\gamma$ is raised to 208, the model accuracy sharply declines to nearly zero.
By inspecting the training logs, we discover that after a certain period of parameter updates in the model training,
the loss value exceeds the upper bound of computer's machine number, resulting in overflow and optimization failure.
When $\gamma$ is set to 256 or 512, the loss value directly overflows before the model parameter updates, preserving the feature extraction capability of the pre-trained model.

For ResNet50 trained with $L_{\text{soft-nu}}$, as $\gamma$ increases,
its $\mathcal A_{\text{SW}}$ and $\mathcal A_{\text{Uni}}$ linearly increase from 49.00\% and 18.55\%.
They reach their peak values of 77.40\% and 63.47\% when $\gamma=48$ and $8$, respectively.
Subsequently, with the further increases in $\gamma$, $\mathcal A_{\text{SW}}$ slowly decreases to 76.68\%,
while $\mathcal A_{\text{Uni}}$ decreases more rapidly to 33.18\%.
When $\gamma\in[32, 192]$, the sample-wise accuracy of the model trained with $L_{\text{bce-nu}}$ is consistently slightly higher than that trained with $L_{\text{soft-nu}}$,
and the uniform accuracy is over 20\% higher during this interval.

The above results about $\gamma$ in the normalized losses indicate that,
compared to the normalized BCE, the normalized SoftMax performs more stable with respect to $\gamma$.
For the normalized BCE, it is necessary to set $\gamma$ into an appropriate interval for well training,
but it can achieve better results than the SoftMax when the $\gamma$ is in an appropriate interval.

\begin{figure*}[!tbhp]
	\centering
    \includegraphics*[scale=0.365, viewport=36 10 730 398]{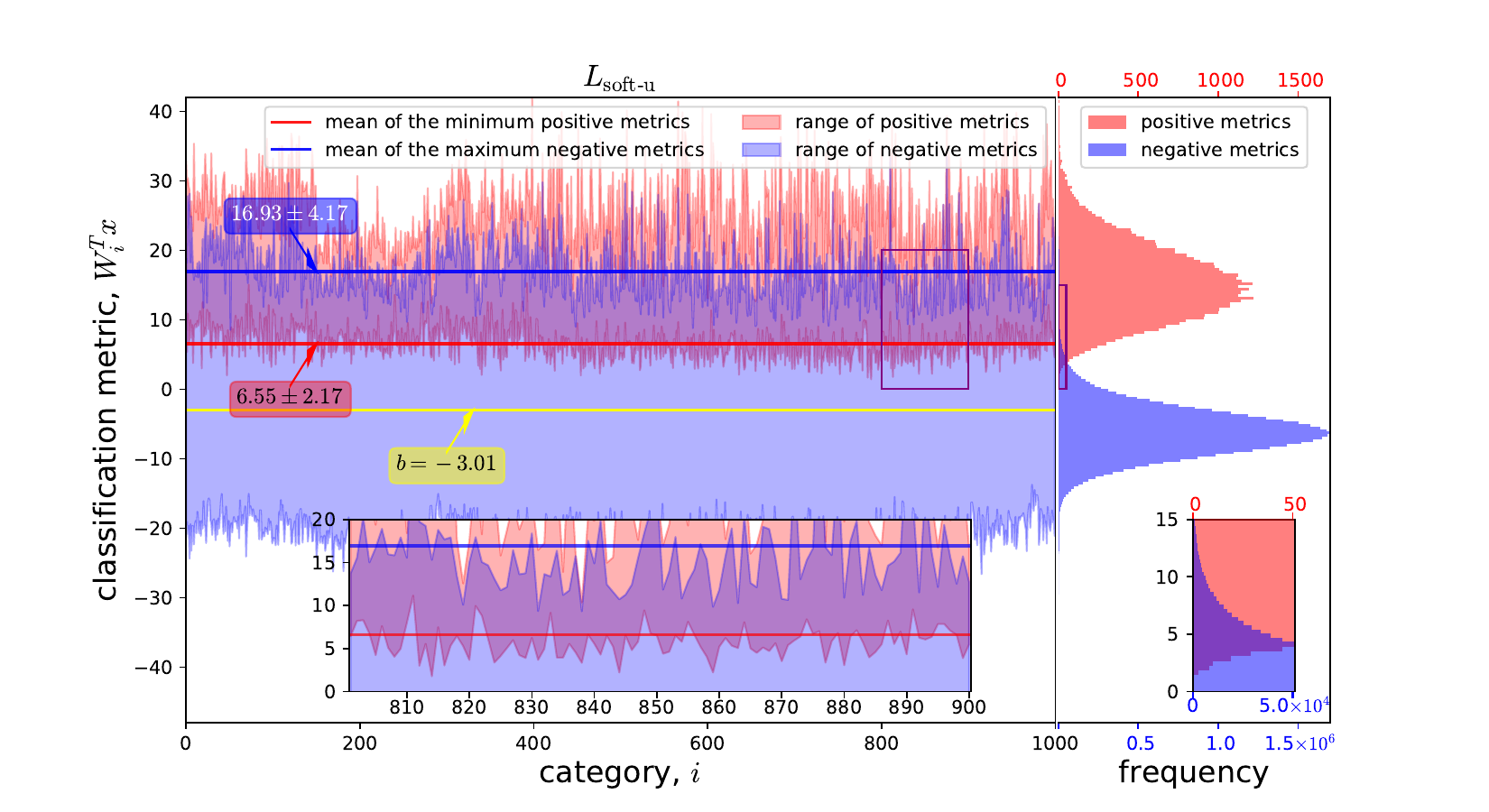}
    \includegraphics*[scale=0.365, viewport=36 10 720 398]{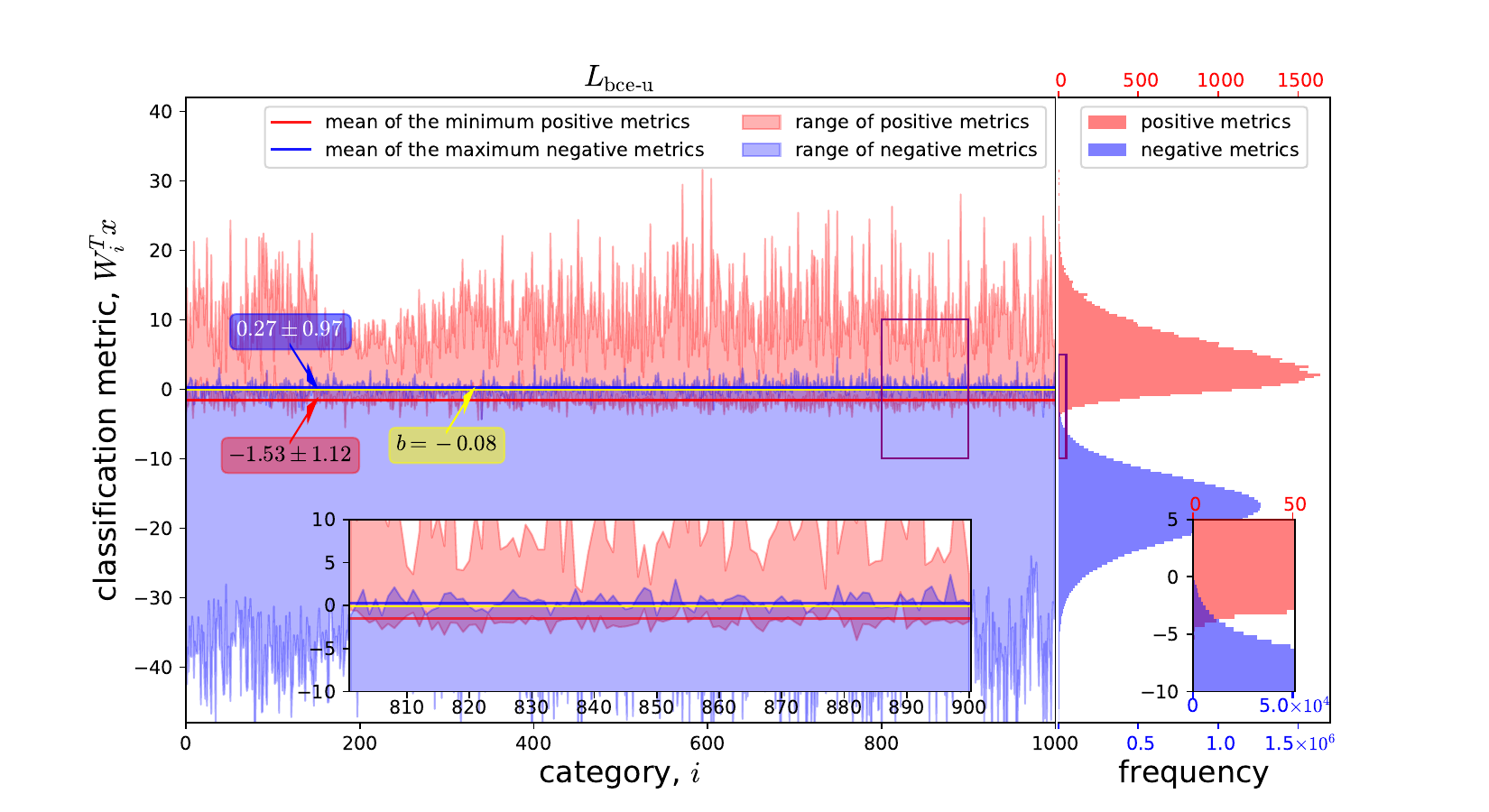}\\\vspace{5pt}
    \includegraphics*[scale=0.365, viewport=36 10 730 398]{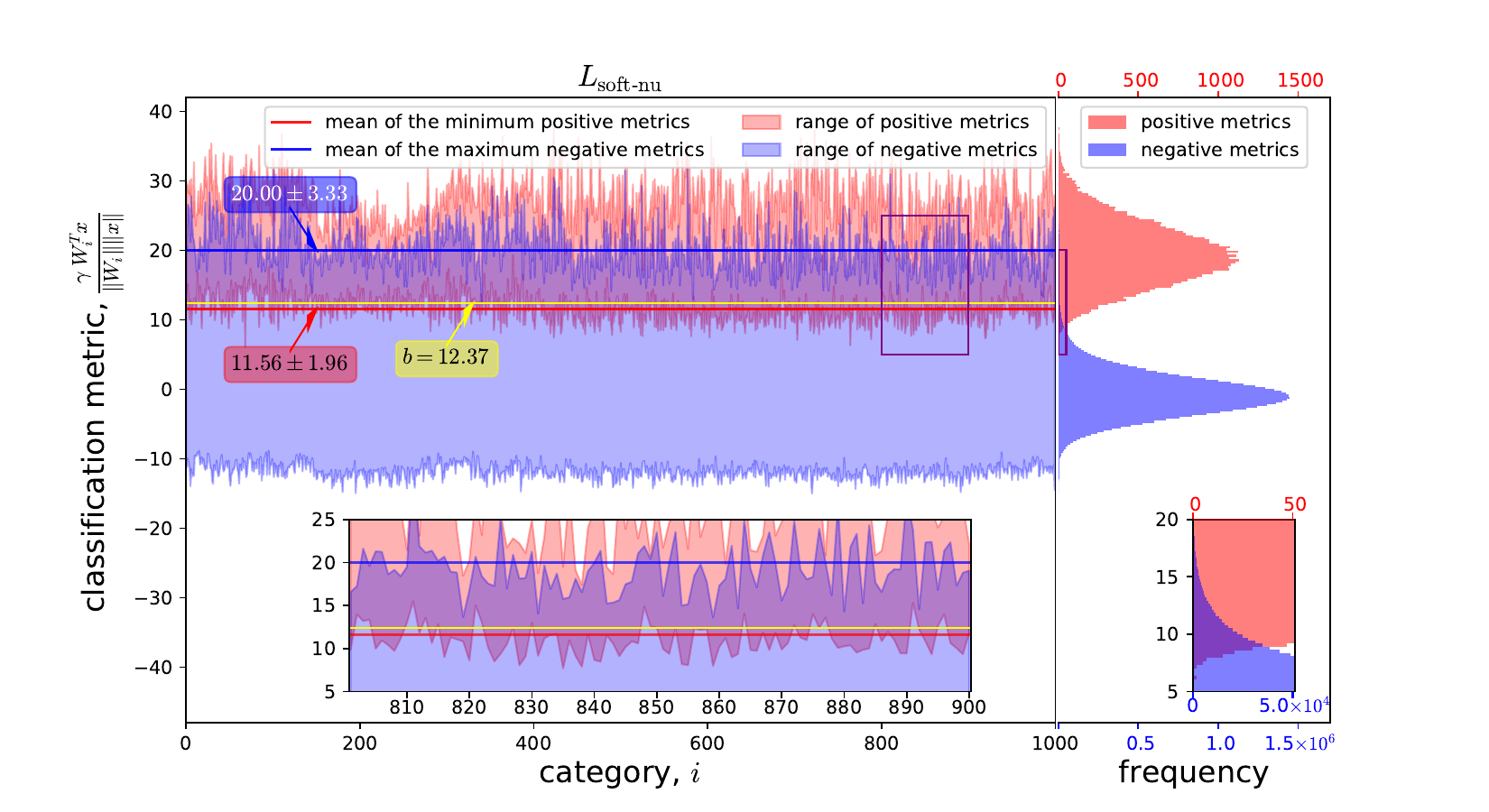}
    \includegraphics*[scale=0.365, viewport=36 10 720 398]{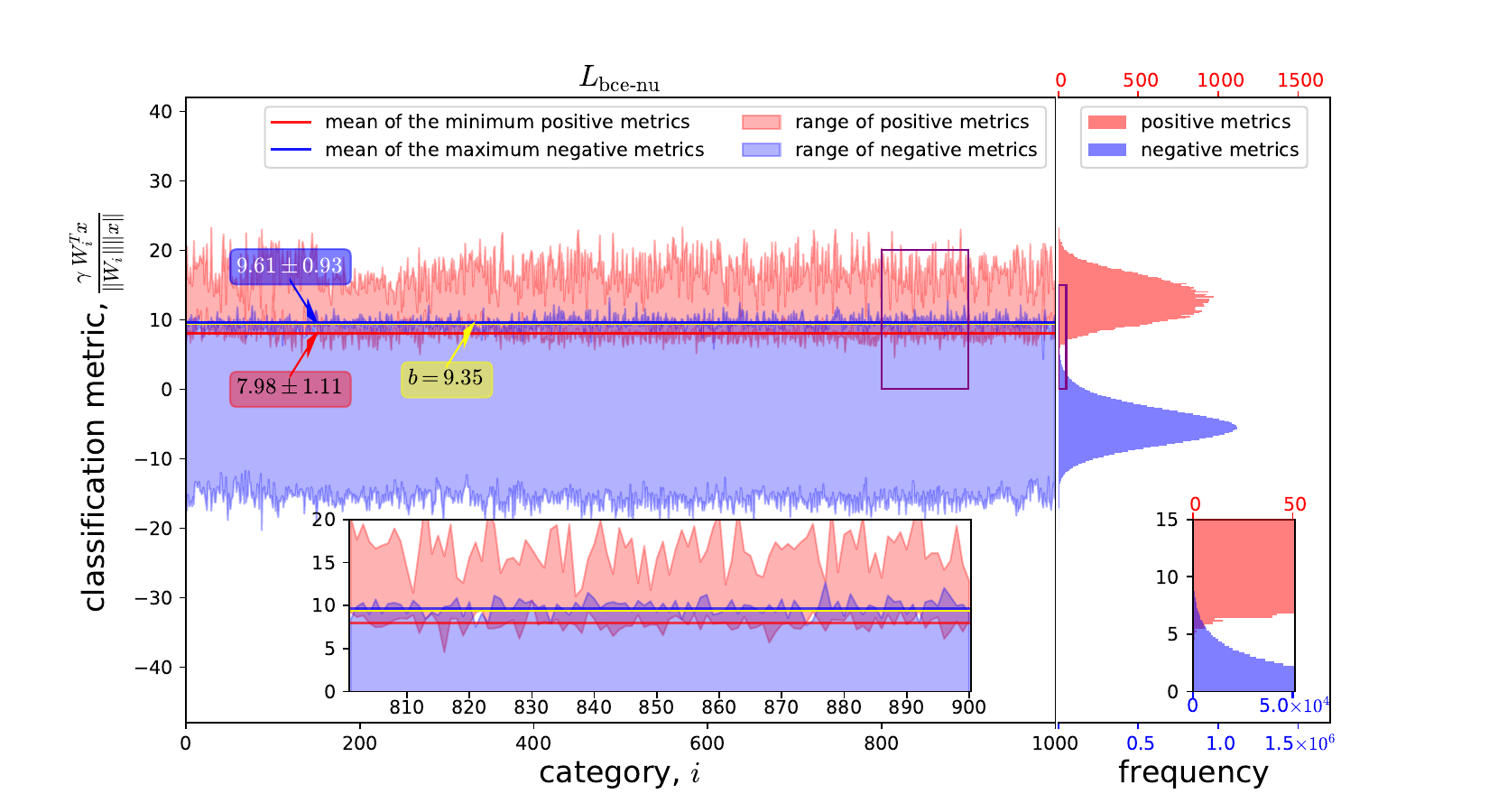}
	\caption{The distributions of positive and negative classification metrics of ResNet50 trained by $L_{\text{soft-nu}}$ (left) and $L_{\text{bce-nu}}$ (right) on ImageNet-1K.
    The smaller overlap between the positive and negative metrics of $L_{\text{bce-u}}$ and $L_{\text{bce-nu}}$
    indicates that the BCE losses are more suitable for uniform classification compared to the SoftMax losses,
    and the their final learned biases (the yellow lines) are closer to the corresponding unified thresholds.
    }
	\label{fig_positive_negative_metric}
\end{figure*}

The \textbf{positive and negative classification metrics} derived by different loss functions exhibit significantly different distribution.
For the ResNet50 trained with $L_{\text{soft-u}}$, $L_{\text{soft-nu}}$, $L_{\text{bce-u}}$, and $L_{\text{bce-nu}}$.
Fig. \ref{fig_positive_negative_metric} illustrates the distributions of
positive and negative classification metrics (without the bias) for the samples correctly classified by them on the validation set.
In the left part of each sub-figure, the light red and light blue regions respectively mark the ranges of positive and negative metrics for each class.
The overlap range of positive and negative metrics for classes 801 to 900, highlighted by the purple boxes, is magnified for better illustration.
The red line and blue line indicate the mean of the minimum positive metrics and that of the maximum negative metrics for all classes.
Histograms of positive and negative metrics are shown in the right part of each sub-figure, with an enlarged view for the overlapping portion highlighted by the purple box.

The uniform classification requires Eq. (\ref{eq_uniform_classification_condition}) to hold,
meaning that the minimum positive metric should be greater than the maximum negative metric.
It is challenging to achieve this ideal state in the experiments, leading to the overlap between the positive and negative metric regions.
Nevertheless, it is evident from the figure that the models trained with the BCE losses
are closer to the state described by the Eq. (\ref{eq_uniform_classification_condition}),
 compared to the models trained with the SoftMax losses.

Firstly, compared to the SoftMax losses, the overlap regions corresponding to the BCE losses are very small.
For example, the overlap region for the model trained with $L_{\text{soft-u}}$ is approximately [6.55, 16.93] with a width of 10.38,
while that for $L_{\text{bce-u}}$ is approximately [$-$1.53, 0.27] with a width of about 1.80 only.
When using the normalized losses, the respective overlap region widths are reduced to 8.44 and 1.63.
Their histograms also align with the above results.
Secondly, compared to the SoftMax losses, the means of minimum positive metrics and maximum negative metrics for each class corresponding to the BCE loss function are more concentrated.
For instance, when training the model with the $L_{\text{soft-u}}$, the means' standard deviations are 4.17 and 2.17, respectively,
while for $L_{\text{bce-u}}$, they are 0.97 and 1.12.
When using the normalized losses, the variances are further reduced.
These results suggest that the BCE loss more uniformly distinguishes the positive and negative metric for the all samples.
In addition, the yellow lines in the figures denote the bias values learned by the different classifiers after the training.
It can be observed that the biases learned by $L_{\text{bce-u}}$ and $L_{\text{bce-nu}}$ are $-$0.08 and 9.35,
more effectively distinguishing the positive and negative metrics of the all samples.
According to Table \ref{Tab_imagenet1k_1}, the optimal thresholds for the metrics without bias can be calculated as $-$0.318 and 9.154, respectively,
which are relatively close to the learned biases, indicating that the BCE losses can learn the optimal threshold.
The SoftMax function does not exhibit this ability in its results.

Furthermore, the positive metrics of $L_{\text{bce-u}}$ and $L_{\text{bce-nu}}$ distribute in a narrower range, compared with that of $L_{\text{soft-u}}$ and $L_{\text{soft-nu}}$,
indicating stronger intra-class compactness for the features extracted by the models trained by the BCE losses.
The smaller overlap of positive and negative metrics associated with the BCE losses
suggests the smaller intersection of feature hyperspheres of different categories,
thereby indicating the higher inter-class distinctiveness.

\begin{table}[tbph]
  \centering
  \caption{The statistic information about the FGVC datasets}\label{Tab_FGCV-statistic}
  \setlength{\tabcolsep}{1.50mm}{
\begin{tabular}{c|c|ccccc}
  \hline
            \multicolumn{2}{c|}{Dataset}& CUB  & Aircraft & Flowers102 & Cars & Dogs \\\hline
  \multirow{2}{*}{Image No.}&Training  & 5994 & 6667 & 2040 & 8144 & 12000 \\
  &Testing      & 5794 & 3333 & 6149 & 8041 & 8580  \\\hline
  \multicolumn{2}{c|}{Class No.}     & 200  & 100  & 102  & 196  & 120   \\\hline
\end{tabular}}
\end{table}

\begin{table*}[!tbph]
  \centering
  \caption{The classification performances of twelve loss functions on CUB}\label{Tab_CUB_200_2011}
  \setlength{\tabcolsep}{1.00mm}{
  \begin{tabular}{c|c|| ccc| ccc| ccc| ccc}\hline
  Model & Accuracy &$L_{\text{soft-0}}$    &\makecell[c]{$L_{\text{soft-d}}$\\(baseline)}    &$L_{\text{soft-u}}$    &$L_{\text{soft-n0}}$    &$L_{\text{soft-nd}}$    &$L_{\text{soft-nu}}$
                                    &$L_{\text{bce-0}}$    &$L_{\text{bce-d}}$    &$L_{\text{bce-u}}$    &$L_{\text{bce-n0}}$    &$L_{\text{bce-nd}}$    &$L_{\text{bce-nu}}$\\\hline\hline

\multirow{4}{*}{ResNet50} &$\mathcal A_{\text{SW}}$  &81.98    &81.86    &81.84    &84.88    &84.93    &84.93    &79.46    &79.39    &79.27    &84.00    &85.48    &\textbf{86.09}\\
&$\mathcal A_{\text{CW}}$  &65.50    &65.67    &65.65    &74.53    &74.37    &74.47    &73.75    &73.89    &73.73    &79.44    &82.07    &\textbf{82.46}\\
&$\mathcal A_{\text{Uni}}$ &58.37    &58.03    &57.90    &68.07    &69.07    &69.30    &70.40    &70.62    &70.35    &76.13    &79.58    &\textbf{80.12}\\
&$t^*$      &5.804   &6.068   &5.085   &14.689   &6.842   &6.337   &$-$1.789   &$-$1.614   &$-$1.817   &$-$1.998   &$-$1.859   &$-$2.423\\\hline

\multirow{4}{*}{ResNet101} &$\mathcal A_{\text{SW}}$  &82.97    &83.24    &83.48    &86.33    &85.57    &85.57    &79.44    &79.82    &80.45    &85.85    &\textbf{86.83}    &86.28\\
&$\mathcal A_{\text{CW}}$  &70.28    &70.00    &70.61    &76.82    &76.56    &76.39    &76.51    &76.75    &77.17    &83.19    &83.12    &\textbf{83.33}\\
&$\mathcal A_{\text{Uni}}$ &64.91    &64.46    &64.62    &71.64    &71.56    &71.42    &74.09    &74.25    &74.80    &80.57    &\textbf{80.93}    &80.82\\
&$t^*$      &5.975   &5.855   &5.445   &12.335   &4.605   &4.684   &$-$1.398   &$-$1.486   &$-$1.169   &$-$1.776   &$-$1.343   &$-$2.161\\\hline

\multirow{4}{*}{DenseNet161} &$\mathcal A_{\text{SW}}$&83.79    &83.85    &84.04    &86.97    &86.78    &86.37    &77.93    &77.70    &77.99    &\textbf{87.47}    &87.11    &87.25\\
&$\mathcal A_{\text{CW}}$  &78.15    &77.61    &78.15    &77.70    &78.01    &76.61    &75.20    &75.34    &75.27    &81.95    &81.72    &\textbf{82.05}\\
&$\mathcal A_{\text{Uni}}$ &74.61    &73.66    &74.54    &73.40    &73.99    &71.97    &72.44    &72.83    &72.71    &78.46    &78.81    &\textbf{78.96}\\
&$t^*$      &3.106   &2.699   &2.738   &9.279   &1.803   &1.417   &$-$1.114   &$-$1.187   &$-$1.344   &$-$2.971   &$-$2.309   &$-$2.114\\\hline
  \end{tabular}}
\end{table*}
\begin{table*}[!tbph]
  \centering
  \caption{The classification performances of twelve loss functions on Aircraft}\label{Tab_FGVC-aircraft}
  \setlength{\tabcolsep}{1.00mm}{
  \begin{tabular}{c|c|| ccc| ccc| ccc| ccc}\hline
  Model & Accuracy &$L_{\text{soft-0}}$    &\makecell[c]{$L_{\text{soft-d}}$\\(baseline)}    &$L_{\text{soft-u}}$    &$L_{\text{soft-n0}}$    &$L_{\text{soft-nd}}$    &$L_{\text{soft-nu}}$
                                    &$L_{\text{bce-0}}$    &$L_{\text{bce-d}}$    &$L_{\text{bce-u}}$    &$L_{\text{bce-n0}}$    &$L_{\text{bce-nd}}$    &$L_{\text{bce-nu}}$\\\hline\hline

\multirow{4}{*}{ResNet50} &$\mathcal A_{\text{SW}}$  &91.42    &91.48    &91.39    &91.33    &91.24    &91.39    &90.52    &90.46    &90.73    &90.82    &92.11    &\textbf{92.14}\\
&$\mathcal A_{\text{CW}}$  &82.87    &81.07    &80.71    &81.94    &82.51    &82.42    &89.20    &88.84    &89.29    &88.48    &\textbf{90.19}    &90.13\\
&$\mathcal A_{\text{Uni}}$ &76.09    &74.41    &73.78    &76.18    &76.03    &77.74    &87.67    &87.16    &87.88    &86.38    &88.42    &\textbf{88.45}\\
&$t^*$      &8.653   &8.899   &7.994   &17.554   &11.165   &10.441   &$-$0.354   &$-$0.698   &$-$0.301   &$-$0.930   &$-$0.725   &$-$1.330\\\hline

\multirow{4}{*}{ResNet101} &$\mathcal A_{\text{SW}}$  &92.11    &91.87    &92.05    &91.90    &92.05    &91.87    &90.82    &91.06    &90.85    &91.51    &92.08    &\textbf{92.41}\\
&$\mathcal A_{\text{CW}}$  &86.65    &85.78    &86.62    &84.73    &85.57    &85.24    &90.04    &90.01    &89.95    &90.16    &90.73    &\textbf{91.18}\\
&$\mathcal A_{\text{Uni}}$ &82.51    &81.61    &82.87    &80.38    &82.03    &81.91    &88.66    &88.84    &88.87    &88.66    &89.47    &\textbf{89.59}\\
&$t^*$      &8.530   &8.548   &8.145   &14.920   &8.296   &8.065   &$-$0.688   &$-$0.248   &$-$0.245   &$-$0.164   &$-$1.151   &$-$1.231\\\hline

\multirow{4}{*}{DenseNet161} &$\mathcal A_{\text{SW}}$ &91.99    &92.20    &\textbf{92.71}    &91.90    &91.21    &91.63    &91.33    &91.24    &91.36    &92.44    &92.44    &92.38\\
&$\mathcal A_{\text{CW}}$  &89.08    &89.08    &89.35    &85.06    &83.92    &84.58    &90.76    &90.64    &\textbf{90.97}    &89.71    &89.41    &89.47\\
&$\mathcal A_{\text{Uni}}$ &86.74    &86.20    &86.92    &81.97    &80.53    &80.59    &89.53    &89.17    &\textbf{89.62}    &87.88    &87.07    &87.49\\
&$t^*$      &4.677   &3.489   &3.711   &12.168   &4.907   &4.101   &$-$0.043   &$-$0.164   &$-$0.712   &$-$2.533   &$-$1.178   &$-$1.958\\\hline
  \end{tabular}}
\end{table*}
\begin{table*}[!tbph]
  \centering
  \caption{The classification performances of twelve loss functions on Flowers102}\label{Tab_Flowers102}
  \setlength{\tabcolsep}{1.00mm}{
  \begin{tabular}{c|c|| ccc| ccc| ccc| ccc}\hline
  Model & Accuracy &$L_{\text{soft-0}}$    &\makecell[c]{$L_{\text{soft-d}}$\\(baseline)}    &$L_{\text{soft-u}}$    &$L_{\text{soft-n0}}$    &$L_{\text{soft-nd}}$    &$L_{\text{soft-nu}}$
                                    &$L_{\text{bce-0}}$    &$L_{\text{bce-d}}$    &$L_{\text{bce-u}}$    &$L_{\text{bce-n0}}$    &$L_{\text{bce-nd}}$    &$L_{\text{bce-nu}}$\\\hline\hline

\multirow{4}{*}{ResNet50} &$\mathcal A_{\text{SW}}$ &95.33    &95.97    &95.90    &96.76    &96.94    &96.28    &95.58    &95.51    &95.41    &\textbf{97.20}    &96.91    &97.15\\
&$\mathcal A_{\text{CW}}$  &87.98    &89.20    &88.97    &91.35    &92.06    &91.36    &93.36    &93.45    &92.67    &95.06    &\textbf{95.23}    &94.89\\
&$\mathcal A_{\text{Uni}}$ &85.18    &86.83    &85.88    &88.88    &89.90    &89.12    &92.21    &92.45    &91.58    &93.85    &\textbf{94.24}    &93.97\\
&$t^*$      &7.545   &7.166   &6.854   &13.558   &6.222   &6.329   &$-$1.552   &$-$1.712   &$-$1.564   &$-$1.501   &$-$2.023   &$-$1.945\\\hline

\multirow{4}{*}{ResNet101} &$\mathcal A_{\text{SW}}$  &96.23    &96.52    &96.71    &96.93    &97.27    &97.01    &95.87    &96.06    &96.26    &97.67    &\textbf{97.79}    &97.66\\
&$\mathcal A_{\text{CW}}$  &91.77    &92.08    &92.52    &93.25    &92.97    &91.82    &94.50    &94.76    &94.86    &95.95    &\textbf{96.18}    &96.02\\
&$\mathcal A_{\text{Uni}}$ &89.72    &89.93    &90.71    &91.14    &90.94    &89.35    &93.62    &93.90    &94.00    &95.04    &95.20    &\textbf{95.23}\\
&$t^*$      &6.756   &6.237   &6.653   &11.201   &4.207   &3.831   &$-$1.694   &$-$1.321   &$-$1.368   &$-$1.948   &$-$1.416   &$-$1.056\\\hline

\multirow{4}{*}{DenseNet161} &$\mathcal A_{\text{SW}}$ &96.96    &96.58    &96.81    &96.76    &96.73    &96.80    &96.85    &97.09    &96.08    &\textbf{98.00}    &97.76    &97.69\\
&$\mathcal A_{\text{CW}}$  &94.97    &94.65    &95.09    &91.56    &91.95    &92.21    &96.10    &\textbf{96.16}    &96.05    &94.63    &94.96    &94.88\\
&$\mathcal A_{\text{Uni}}$ &93.93    &93.46    &94.05    &89.15    &89.88    &89.77    &95.35    &\textbf{95.54}    &95.50    &93.12    &93.72    &93.66\\
&$t^*$      &4.166   &3.867   &3.980   &8.092   &1.010   &0.864   &$-$2.498   &$-$1.931   &$-$1.794   &$-$3.688   &$-$3.611   &$-$3.426\\\hline
  \end{tabular}}
\end{table*}
\begin{table*}[!tbph]
  \centering
  \caption{The classification performances of twelve loss functions on Cars}\label{Tab_Stanford_Cars}
  \setlength{\tabcolsep}{1.00mm}{
  \begin{tabular}{c|c|| ccc| ccc| ccc| ccc}\hline
  Model & Accuracy &$L_{\text{soft-0}}$    &\makecell[c]{$L_{\text{soft-d}}$\\(baseline)}    &$L_{\text{soft-u}}$    &$L_{\text{soft-n0}}$    &$L_{\text{soft-nd}}$    &$L_{\text{soft-nu}}$
                                    &$L_{\text{bce-0}}$    &$L_{\text{bce-d}}$    &$L_{\text{bce-u}}$    &$L_{\text{bce-n0}}$    &$L_{\text{bce-nd}}$    &$L_{\text{bce-nu}}$\\\hline\hline

\multirow{4}{*}{ResNet50} &$\mathcal A_{\text{SW}}$ &93.21    &93.22    &93.16    &93.68    &93.36    &93.28    &92.87    &92.59    &92.44    &92.65    &\textbf{93.91}    &93.86\\
&$\mathcal A_{\text{CW}}$  &82.45    &81.82    &82.35    &86.21    &86.23    &85.75    &90.87    &90.30    &90.35    &89.03    &\textbf{92.02}    &91.94\\
&$\mathcal A_{\text{Uni}}$ &76.48    &75.84    &76.20    &81.18    &81.66    &81.32    &88.92    &88.20    &88.14    &86.93    &90.23    &\textbf{90.25}\\
&$t^*$      &8.652   &8.442   &8.290   &15.380   &7.687   &7.447   &$-$1.348   &$-$1.231   &$-$1.530   &$-$1.586   &$-$1.328   &$-$1.796\\\hline

\multirow{4}{*}{ResNet101} &$\mathcal A_{\text{SW}}$ &93.30    &93.51    &93.47    &93.57    &93.53    &93.62    &92.76    &92.30    &92.58    &93.69    &\textbf{94.23}    &94.14\\
&$\mathcal A_{\text{CW}}$  &85.46    &85.37    &85.92    &86.82    &86.95    &87.27    &91.61    &90.96    &91.31    &91.84    &\textbf{92.60}    &92.35\\
&$\mathcal A_{\text{Uni}}$ &80.25    &80.06    &81.00    &81.93    &82.88    &82.61    &90.05    &89.35    &89.73    &89.69    &\textbf{90.75}    &90.71\\
&$t^*$      &8.608   &8.164   &7.060   &13.350   &5.241   &4.527   &$-$0.876   &$-$0.574   &$-$0.717   &$-$0.674   &$-$0.795   &$-$0.794\\\hline

\multirow{4}{*}{DenseNet161} &$\mathcal A_{\text{SW}}$ &92.96    &92.96    &93.32    &92.55    &92.35    &92.75    &91.79    &91.46    &91.43    &\textbf{93.63}    &93.22    &93.53\\
&$\mathcal A_{\text{CW}}$  &88.98    &88.58    &88.88    &84.55    &84.68    &84.58    &\textbf{90.71}    &90.27    &90.32    &90.23    &89.88    &90.05\\
&$\mathcal A_{\text{Uni}}$ &85.72    &85.29    &85.47    &80.29    &80.24    &80.20    &\textbf{88.93}    &88.53    &88.53    &87.78    &87.22    &87.59\\
&$t^*$      &3.109   &3.428   &3.347   &7.621   &0.742   &$-$0.266   &$-$0.207   &$-$0.641   &$-$0.250   &$-$2.999   &$-$2.387   &$-$3.003\\\hline
  \end{tabular}}
\end{table*}
\begin{table*}[!tbph]
  \centering
  \caption{The classification performances of twelve loss functions on Dogs}\label{Tab_Stanford_Dogs}
  \setlength{\tabcolsep}{1.00mm}{
  \begin{tabular}{c|c|| ccc| ccc| ccc| ccc}\hline
  Model & Accuracy &$L_{\text{soft-0}}$    &\makecell[c]{$L_{\text{soft-d}}$\\(baseline)}    &$L_{\text{soft-u}}$    &$L_{\text{soft-n0}}$    &$L_{\text{soft-nd}}$    &$L_{\text{soft-nu}}$
                                    &$L_{\text{bce-0}}$    &$L_{\text{bce-d}}$    &$L_{\text{bce-u}}$    &$L_{\text{bce-n0}}$    &$L_{\text{bce-nd}}$    &$L_{\text{bce-nu}}$\\\hline\hline

\multirow{4}{*}{ResNet50} &$\mathcal A_{\text{SW}}$  &81.82    &81.83    &81.20    &82.60    &82.70    &82.66    &81.13    &81.26    &81.35    &82.45    &83.11    &\textbf{83.18}\\
&$\mathcal A_{\text{CW}}$  &58.67    &58.02    &58.33    &64.16    &64.74    &64.41    &75.66    &76.14    &76.48    &77.89    &78.15    &\textbf{78.61}\\
&$\mathcal A_{\text{Uni}}$ &54.07    &53.29    &53.89    &60.16    &60.78    &60.06    &74.15    &74.31    &74.84    &76.19    &76.48    &\textbf{76.88}\\
&$t^*$      &6.713   &6.289   &5.471   &16.548   &9.310   &9.108   &$-$1.348   &$-$1.760   &$-$1.634   &$-$1.186   &$-$1.305   &$-$1.268\\\hline

\multirow{4}{*}{ResNet101} &$\mathcal A_{\text{SW}}$ &83.86    &83.53    &83.39    &84.23    &84.17    &83.76    &83.24    &82.48    &82.63    &83.28    &\textbf{84.31}    &83.94\\
&$\mathcal A_{\text{CW}}$  &66.53    &66.57    &65.48    &67.34    &66.60    &66.93    &79.55    &79.02    &78.61    &79.24    &\textbf{80.48}    &79.91\\
&$\mathcal A_{\text{Uni}}$ &62.87    &62.33    &61.61    &63.64    &62.54    &63.31    &77.82    &77.41    &77.19    &77.37    &\textbf{78.72}    &78.25\\
&$t^*$      &6.204   &5.700   &4.930   &14.173   &7.636   &6.168   &$-$1.152   &$-$0.850   &$-$1.017   &$-$1.077   &$-$1.345   &$-$1.298\\\hline

\multirow{4}{*}{DenseNet161} &$\mathcal A_{\text{SW}}$ &84.20    &84.53    &84.71    &85.05    &84.92    &84.91    &82.68    &82.67    &82.63    &84.84    &84.92    &\textbf{85.05}\\
&$\mathcal A_{\text{CW}}$  &75.51    &76.45    &76.12    &70.85    &71.21    &70.08    &80.58    &80.36    &\textbf{80.58}    &79.73    &76.22    &79.71\\
&$\mathcal A_{\text{Uni}}$ &72.90    &73.92    &73.74    &67.82    &68.43    &66.86    &79.17    &79.16    &\textbf{79.21}    &77.77    &73.55    &77.88\\
&$t^*$      &1.732   &1.964   &1.846   &9.127   &2.105   &0.889   &$-$0.278   &$-$0.594   &$-$0.406   &$-$1.487   &$-$0.156   &$-$0.796\\\hline
  \end{tabular}}
\end{table*}
\subsection{Fine-grained visual classification}
\begin{figure*}[!tbhp]
	\centering
    \includegraphics*[scale=0.375, viewport=18 3 640 335]{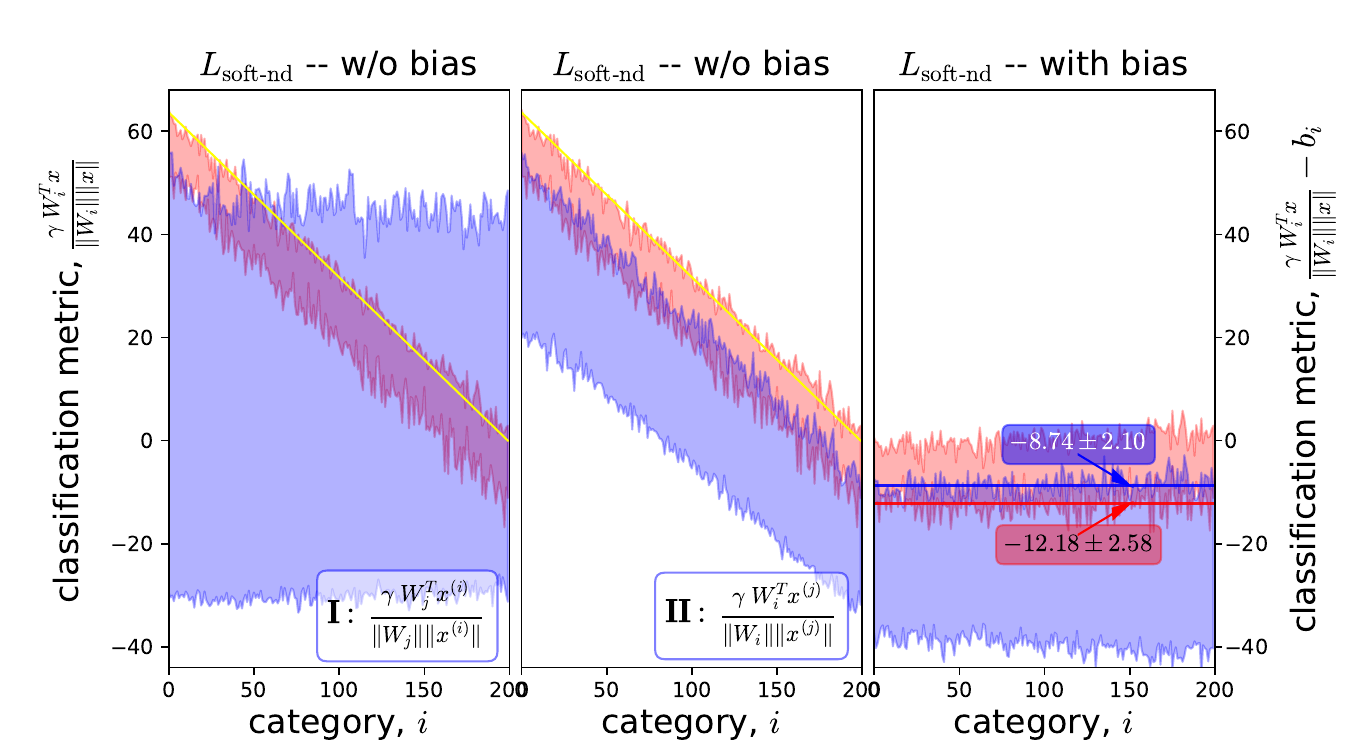}\hspace{15pt}
    \includegraphics*[scale=0.375, viewport=18 3 640 335]{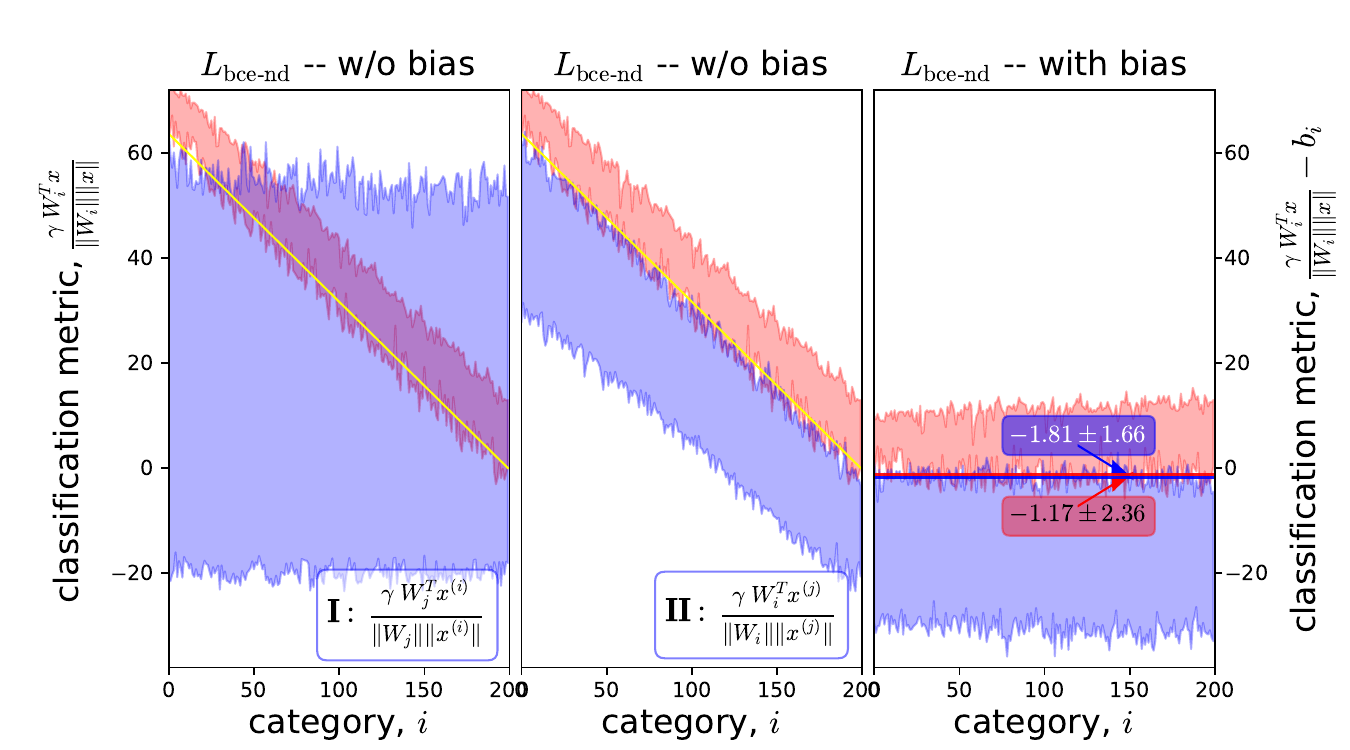}\\\vspace{15pt}
    \includegraphics*[scale=0.375, viewport=18 3 640 335]{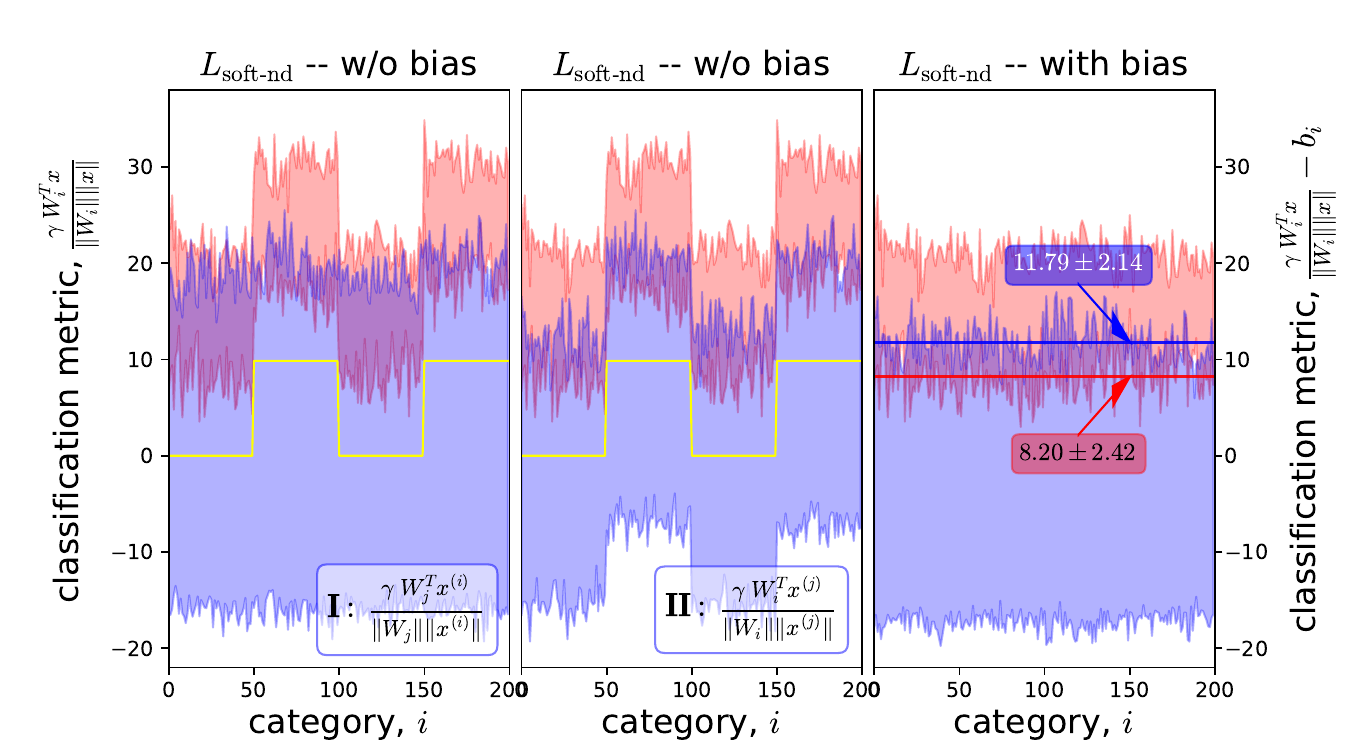}\hspace{15pt}
    \includegraphics*[scale=0.375, viewport=18 3 640 335]{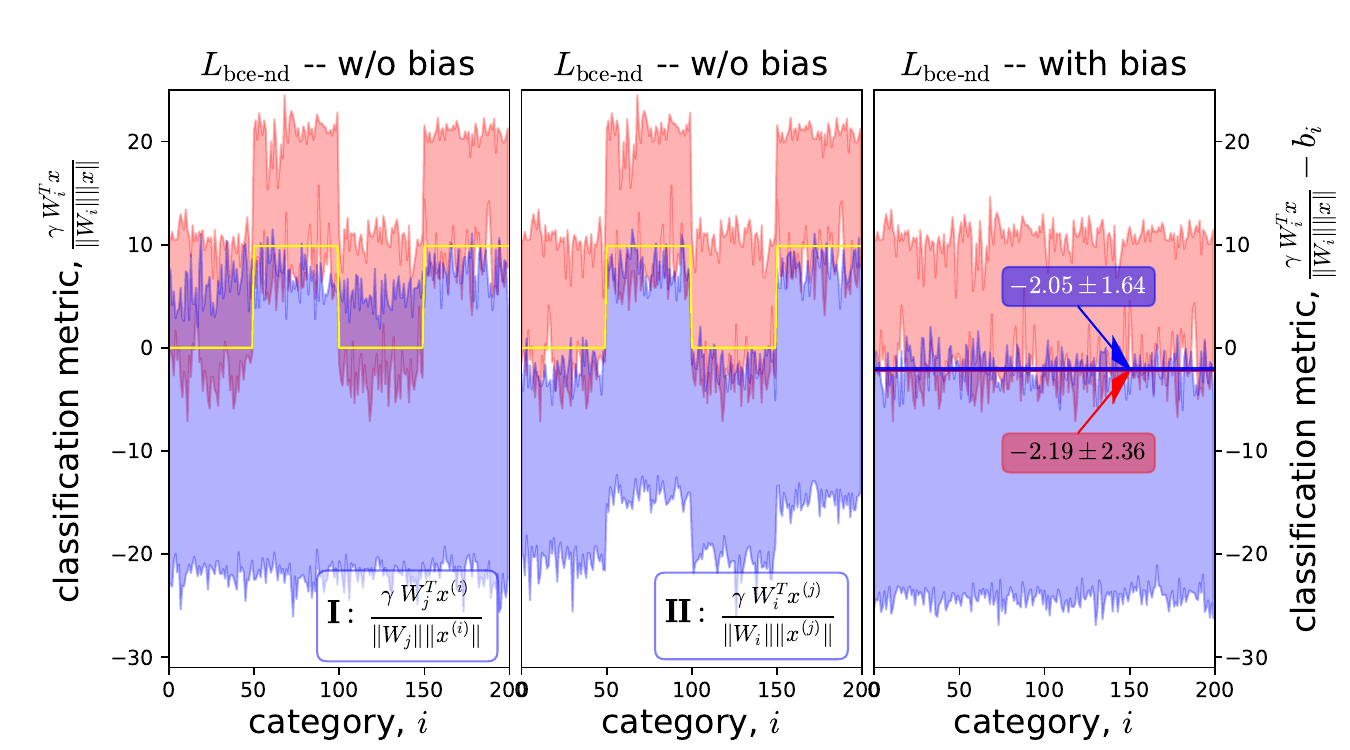}
	\caption{The distributions of positive and negative classification metrics of ResNet50 trained
        by $L_{\text{soft-nd}}$ (left) and $L_{\text{bce-nd}}$ (right) on CUB with bias initialization 3 (top) and 7 (bottom).
        The bias learned by $L_{\text{soft-nd}}$ cannot distinguish the positive and negative metrics,
        indicating it unsuitable for (class-wise) uniform classification.
        Despite $L_{\text{bce-nd}}$ is not suitable for them either,
        its learned biases (the yellow curve) effectively distinguishes the positive and type II negative metrics,
        and, integrated with these biases, it presents more uniformity.
    }
	\label{fig_positive_negative_metric-CUB}
\end{figure*}

To further demonstrate the superiority of the BCE loss over the SoftMax loss,
we train the three models using the twelve losses on five commonly used fine-grained visual classification (FGVC) datasets,
i.e., CUB \cite{wah2011caltech}, Aircraft \cite{maji2013fine}, Flowers102 \cite{nilsback2008automated}, Cars \cite{krause20133d}, and Dogs \cite{khosla2011novel}.
Table \ref{Tab_FGCV-statistic} presents the information about their data size, number of classes, etc.
On these five datasets, we respectively train the three deep networks, i.e., ResNet50, ResNet101, and DenseNet161,
while the models have been pre-trained on ImageNet-1K with 90 epoches.
In each training, we set the initial learning rate to $10^{-3}$ and decay it using a cosine strategy;
we retrain the model for 100 epochs on the training set.
For the normalized SoftMax and BCE losses, we set $\gamma=96$.
Table \ref{Tab_CUB_200_2011} to \ref{Tab_Stanford_Dogs} present the classification results on the testing sets.

Unlike the results on the ImageNet, the normalized BCE loss integrated with unified threshold, $L_{\text{bce-nu}}$, does not always achieve the optimal results in these five tasks.
However, one can still observe some consistent findings.
Firstly, except for the sample-wise accuracy on Aircraft achieved by DensNet161, the all optimal accuracies on the remaining tasks are obtained by using BCE.
In other words, for these classification tasks, \emph{the BCE loss indeed outperforms the SoftMax loss}.
Secondly, after training a model on a given dataset, the optimal sample-wise accuracy and optimal uniform accuracy are usually achieved by different losses.
This phenomenon has been observed in the experiments on the ImageNet,
which suggests that the sample-wise accuracy differ to the uniform accuracy, indicating a distinction between the classification and uniform classification.

By the analysis in Sec. \ref{sec_loss_function}, the diverse biases in the $L_{\text{bce-d}}$ and $L_{\text{bce-nd}}$ play the theoretical roles
in separating the positive and negative metrics for each class.
However, in the above experiments, we do not observe definitive results supporting this viewpoint.
Comparing the values of $b_i$ before and after the training, we find that the update magnitude of the $b_i$ is relatively very small,
which leads us to speculate that the initialization of $b_i$ might affect model performance.
To investigate this hypothesis, we employ various bias initializations.

\begin{table*}[!tbph]
  \centering
  \caption{The performances of diverse bias integrated losses on CUB, with different initializations}\label{Tab_initialization_bias}
  \setlength{\tabcolsep}{1.00mm}{
  \begin{tabular}{c|c ccc cccc | cccc cccc}\hline
    & \multicolumn{8}{c|}{$L_{\text{soft-d}}$}& \multicolumn{8}{c}{$L_{\text{soft-nd}}$}\\\hline
Mode & 0& 1& 2& 3& 4& 5& 6& 7& 0& 1& 2& 3& 4& 5& 6& 7\\\hline
$\mathcal A_{\text{SW}}$  & 81.86 &62.46    &77.34    &77.18    &\textbf{82.21}    &81.96    &81.55    &81.81 & 84.93 &80.55    &84.97    &85.19    &84.98    &84.54    &84.54    &\textbf{85.21}\\
$\mathcal A_{\text{CW}}$  & 65.67 &27.65    &48.64    &45.60    &65.34    &\textbf{65.71}    &62.60    &63.12 & 74.37 &58.28    &\textbf{75.66}    &75.20    &74.28    &74.70    &74.30    &74.89\\
$\mathcal A_{\text{Uni}}$ & \textbf{58.03} &14.29    &32.79    &27.27    &57.80    &57.68    &54.50    &54.95 & 69.07 &26.27    &\textbf{69.93}    &68.88    &69.05    &69.26    &68.95    &69.49\\
$t^*$  &6.07   &137.75   &48.87   &48.62   &15.09   &15.08   &13.90   &13.70 &6.84  &$-$77.06   &$-$9.01   &$-$9.08   &7.44   &6.42   &11.00   &11.59\\\hline

    & \multicolumn{8}{c|}{$L_{\text{bce-d}}$}& \multicolumn{8}{c}{$L_{\text{bce-nd}}$}\\\hline
Mode & 0& 1& 2& 3& 4& 5& 6& 7& 0& 1& 2& 3& 4& 5& 6& 7\\\hline
$\mathcal A_{\text{SW}}$ & \textbf{79.39} &0.52    &0.71    &1.09    &43.80    &49.24    &69.66    &67.35 & \textbf{85.48} &0.52    &84.79    &85.26    &85.42    &85.42    &85.17    &84.38\\
$\mathcal A_{\text{CW}}$ & \textbf{73.89}  &0.50    &0.40    &0.31    &25.63    &30.17    &55.71    &51.43 & 82.07 &0.28    &81.95    &\textbf{82.38}    &81.90    &81.90    &81.50    &80.65\\
$\mathcal A_{\text{Uni}}$ & \textbf{70.62} &0.50    &0.26    &0.10    &17.09    &21.64    &50.12    &45.91 &79.58&0.26    &79.15    &\textbf{80.01}    &79.34    &79.32    &78.68    &78.32\\
$t^*$     &$-$1.61 &198.55   &$-$5.15   &$-$4.48   &$-$1.51   &$-$1.17   &$-$1.06   &$-$1.04 & $-$1.86 &0.62   &$-$1.14   &$-$1.31   &$-$2.34   &$-$1.96   &$-$1.68   &$-$2.00\\\hline
  \end{tabular}}
\end{table*}

\begin{table}
  \centering
  \caption{The mean ($\times10^{-3}$) and standard deviation ($\times10^{-3}$) of the change of bias in $L_{\text{bce-nd}}$ and $L_{\text{soft-nd}}$ with different initialization modes before and after the training.}\label{Tab_bias_change}
  \setlength{\tabcolsep}{0.4mm}{
\begin{tabular}{c|ccccccc}
  \hline
  Mode & 1 & 2 & 3 & 4 & 5 & 6 & 7 \\\hline
  \hline
  $L_{\text{soft-nd}}$ & $-$18.4$\pm$99 & 3.1$\pm$6.5 & 2.9$\pm$6.6
                          & 0.7$\pm$0.9  & 0.5$\pm$0.9  & 0.6$\pm$1.9 & 0.6$\pm$1.9\\
  $L_{\text{bce-nd}}$ & - & $-$6.5$\pm$8.1 & $-$6.8$\pm$8.3 & 2.1$\pm$0.9
                          & 2.1$\pm$1.0 & 2.9$\pm$1.6 & 3.1$\pm$1.5\\\hline
\end{tabular}}
\end{table}
In the diverse bias integrated losses, $L_{\text{soft-d}}$, $L_{\text{soft-nd}}$, $L_{\text{bce-d}}$, and $L_{\text{bce-nd}}$,
the bias default initialization (Mode 0) adopts the random numbers with mean of zero.
We compare it with seven other modes as following,
\begin{itemize}
  \item Mode 1: $b_i = i$ for $1\leq i\leq N$;
  \item Mode 2: $b_i = \frac{64*i}{N}$ for $1\leq i\leq N$;
  \item Mode 3: $b_i = \frac{64*(N-i)}{N}$ for $1\leq i\leq N$;
  \item Mode 4: $b_i = \log(96*i)$ for $1\leq i\leq N$;
  \item Mode 5: $b_i = \log(96*(N+1-i))$ for $1\leq i\leq N$;
  \item Mode 6: $b_i = \log(96*N)$ for $1\leq i\leq N/4$ and $N/2\leq i\leq 3N/4$; $b_i=0$ for others;
  \item Mode 7: $b_i = 0$ for $1\leq i\leq N/4$ and $N/2\leq i\leq 3N/4$; $b_i=\log(96*N)$ for others.
\end{itemize}

Table \ref{Tab_initialization_bias} presents the performances of ResNet50 trained with $L_{\text{soft-d}}$, $L_{\text{soft-nd}}$, $L_{\text{bce-d}}$, and $L_{\text{bce-nd}}$ on the CUB.
According to the table, one can find that (a) compared to the SoftMax losses, the performance of the BCE losses is more sensitive to the initialization mode of bias.
For example, when initializing bias using Mode1/Mode2/Mode3, the model trained with $L_{\text{bce-d}}$ experiences overflow, resulting in optimization failure.
When initializing bias using Mode4/Mode5/Mode6/Mode7, the performance of the model trained with $L_{\text{bce-d}}$ also significantly decreases,
with a reduction of over 10\% in sample-wise accuracy and over 20\% in uniform accuracy.
In contrast, for $L_{\text{soft-d}}$, the model converges regardless of the initialization method,
and it even achieves better sample-wise accuracy when initializing the bias using Mode4.
(b) When using different bias initialization modes,
the performance of $L_{\text{soft-nd}}$ and $L_{\text{bce-nd}}$ is significantly better than that of $L_{\text{soft-d}}$ and $L_{\text{bce-d}}$,
approaching or even surpassing the performance of the default initialization mode.
This indicates that the normalization could decrease the sensitivity of the loss to the bias initialization and thereby increase the training stability.

To further investigate the roles of $b_i$ in the classification,
we first calculate the change of bias before and after the training.
Table \ref{Tab_bias_change} shows the mean and standard deviation of bias in $L_{\text{bce-nd}}$ and $L_{\text{soft-nd}}$ with different initialization modes.
One can find that the change of bias are on the order of $10^{-3}$, indicating very small variations before and after the training.

Fig. \ref{fig_positive_negative_metric-CUB} illustrates the distribution of positive and negative metrics of the CUB testing samples correctly classified by ResNet50
trained with $L_{\text{soft-nd}}$ (left) and  $L_{\text{bce-nd}}$ (right) using initialization modes 3 (top) and 7 (bottom).
In each subfigure, the left side displays the distribution of positive metrics (light red) and type I negative metrics (light blue) without bias,
the middle shows that of positive metrics and type II negative metrics without bias,
while the right side depicts the distribution of positive and negative metrics with biases.
The yellow curve represents the biases after training,
while the red and blue lines denote the mean of the minimum positive metrics and that of the maximum negative metrics with bias across all classes.
From the figure, the following observations can be easily made:
(1) The final bias in the classifier significantly influences the positive and negative metrics.
The distributions of positive metrics and type II negative metrics corresponding to $L_{\text{soft-nd}}$ and $L_{\text{bce-nd}}$ closely follow the trend of the bias.
(2) Compared to the $L_{\text{soft-nd}}$, the biases learned by the $L_{\text{bce-nd}}$ almost separates its positive metrics and type II negative metrics.
However, they cannot clearly distinguish the positive metrics from type I negative metrics,
which aligns with our analysis in Sec. \ref{sec_loss_function}, indicating that $L_{\text{bce-nd}}$ is not a suitable loss function for the class-wise uniform classification.
(3) When integrating the biases, the positive and negative metrics for $L_{\text{soft-nd}}$ still exhibit a noticeable overlap,
while the positive and negative metrics of $L_{\text{bce-nd}}$ have very minimal overlap, almost being separable.
In addition, for $L_{\text{bce-nd}}$, the the minimum and maximum values of positive and negative metrics are very close to their respective optimal thresholds,
namely $-1.31$ and $-2.00$ in table \ref{Tab_initialization_bias}.
These results indicate that the BCE loss is indeed suitable for the uniform classification.

\subsection{Face recognition}
Face recognition involves setting a unified threshold in advance, comparing the feature similarity of two facial images with the threshold,
and determining whether these two images are from the same individual, which is a typical open-set task.
In our earlier work \cite{zhou2023uniface}, we have designed the unified threshold integrated cross-entropy (UCE) loss through other means,
which is referred to as $L_{\text{bce-nd}}$ in this paper.
The extensive experiments have been conducted in \cite{zhou2023uniface}, confirming its superiority for face recognition compared to the commonly used SoftMax loss.
We do not repeat them here.

Facial recognition not only requires the intra-class compactness and inter-class distinctiveness of the facial sample features,
but also the uniformity among different sample features.
According to our analysis, the SoftMax losses fail to provide this kind of uniformity.
The $\mathcal L_{\text{bce-d}}$ loss can only offer the uniformity between the positive metrics and II-type negative metrics for the sample features.
The $\mathcal L_{\text{bce-0}}$ achieves the uniformity around the unified threshold $t=0$,
while $\mathcal L_{\text{bce-u}}$ allows for the adjustment of this unified threshold.
The experimental results demonstrate that the normalization can further enhance this uniformity,
as well as the intra-class compactness and inter-class distinctiveness.
Therefore, the normalized BCE loss function with an integrated unified threshold, i.e., $L_{\text{bce-nu}}$ or UCE, achieves the best facial recognition performance in \cite{zhou2023uniface}.

\section{Discussion}

(1) Based on the analysis and experiments in this paper, we speculate that the loss functions suitable for the class-wise uniform classification are
\begin{align}
\nonumber
\mathcal L_{\text{bce-d}_i}(\bm X^{(i)}) = &\log\Big(1+\e^{b_i-c_i(\bm x^{(i)})}\Big)\\
\label{eq_bce_di_loss}
+ \sum_{j=1\atop j\neq i}^N\log\Big(&1+\e^{c_j(\bm x^{(i)})-b_i}\Big),\quad i=1,2,\cdots,N.
\end{align}
They are $N$ BCE losses with different unified bias $b_i$.
Their implementation is slightly complex in the experiments.

(2) To achieve the uniform classification for the feature extraction model $\mathcal M$ and the classifier $\mathcal C$,
relying solely on the loss function is not sufficient;
it requires the coordination of other training strategies.
Moreover, loss functions suitable for uniform classification are not limited to the unified threshold integrated BCE loss, $\mathcal L_{\text{bce-u}}$.
Different design approaches would lead to other loss suitable for the uniform classification.

(3) In the experiments, we have observed that the biases in $L_{\text{bce-d}}$ and $L_{\text{soft-d}}$ update very slowly during the model training,
while they play a crucial role in the classification.
Therefore, it is necessary to tailor a dedicated update strategy for the biases for better model performance.
Conversely, one can consider how to leverage the properties of biases.

(4) We believe that improving the uniformity of features can also benefit the unsupervised learning and other tasks.

\section{Conclusions}
This paper introduces the concept of uniform classification along with the uniform classification accuracy,
and designs the loss function suitable for it, namely, the unified threshold integrated BCE function, $\mathcal L_{\text{bce-u}}$,
which is essentially a BCE function with a unified bias.
Both the math analysis and experimental results indicate that the unified bias in $\mathcal L_{\text{bce-u}}$
leads to the final learning of the unified threshold for uniform classification across all samples,
and improves the uniformity of the final features.
In addition, the experimental results demonstrate that the biases in the general SoftMax and BCE functions
are also playing a crucial role in classification tasks, which could significantly change the positive and negative classification metrics of samples.

{\small
\bibliographystyle{IEEEtran}
\bibliography{egbib}
}

\twocolumn[\begin{@twocolumnfalse}
    \begin{center}
    \LARGE
    {Supplementary for Uniform Classification}
    \end{center}
\end{@twocolumnfalse}]

\renewcommand\thesection{S.\Roman{section}}
\setcounter{section}{0}
\renewcommand\theequation{S.\arabic{equation}}
\setcounter{equation}{0}
\renewcommand\thetable{S.\Roman{table}}
\setcounter{table}{0}

\section{Classification metric matrix}
For better understanding the sample-wise and uniform classifications, we here design classification metric matrix.

Randomly take $N$ samples $\{\bm X^{(i)}\in\mathcal D_i\}_{i=1}^N$
from the dataset $\mathcal D$.
With the model $\mathcal M$ and classifier $\mathcal C=\{c_i(\theta_i;\cdot)\}_{i=1}^N$, we define a metric matrix
\begin{align}
\mathcal {E} =
\setlength{\arraycolsep}{2pt}
\begin{pmatrix}
c_1(\bm x^{(1)})&c_1(\bm x^{(2)})&\cdots&c_1(\bm x^{(N)})\\
c_2(\bm x^{(1)})&c_2(\bm x^{(2)})&\cdots&c_2(\bm x^{(N)})\\
\vdots      &       \vdots      &       \ddots      &       \vdots\\
c_N(\bm x^{(1)})&c_N(\bm x^{(2)})&\cdots&c_N(\bm x^{(N)})\\
\end{pmatrix},
\end{align}
where $\bm x^{(i)} = \mathcal M(\bm X^{(i)})$ for $\forall~i$.

The classification metric matrix will own the diagonal dominant property, when the $N$ samples are correctly classified by the model and classifier:
\begin{enumerate}[~~~(a)]
  \item If the dataset $\mathcal D$ is sample-wise classified by $\mathcal M$ and $\mathcal C$,
        then the diagonal elements of the metric matrix $\mathcal {E}$ is \textbf{column-dominant}.
  \item If the dataset $\mathcal D$ is uniformly classified by $\mathcal M$ and $\mathcal C$,
        then the diagonal elements of the metric matrix $\mathcal {E}$ are \textbf{global-dominant}, i.e., each of its diagonal elements is greater than any non-diagonal element.
\end{enumerate}

\section{The Four Inequalities}
\label{sec_four_inequalities}
We here present the detailed derivations for the inequalities about the naive loss $\mathcal L_{\text{naive}}$ in Eq. (\ref{eq_naive_loss}).
For any sample $\bm X^{(i)}$ captured from category $i$, with feature $\bm x^{(i)} = \mathcal{M}(\bm X^{(i)})$,
the naive loss $\mathcal L_{\text{naive}}$ comprises of one positive and $N-1$ negative metrics,
\begin{align}
\label{eq_naive_loss_supp}
\mathcal L_{\text{naive}}(\bm X^{(i)}) = & - c_i(\bm x^{(i)})  + \frac{1}{N-1}\sum_{j=1\atop j\neq i}^N c_j(\bm x^{(i)}).
\end{align}

Using the inequality of arithmetic and geometric means, i.e., $\sqrt[n]{\prod_{i=1}^na_i} \leq \frac{1}{n}\sum_{i=1}^na_i$ for $a_i\geq0$, we derive four inequalities about $\mathcal L_{\text{naive}}$.
\begin{align}
\nonumber
&\mathcal L_{\text{naive}}(\bm X^{(i)}) \\
=~&- c_i(\bm x^{(i)})  + \frac{1}{N-1}\sum_{j=1\atop j\neq i}^N c_j(\bm x^{(i)})\\
=~&-\frac{N}{N-1}\Big(\frac{N-1}{N}c_i(\bm x^{(i)}) - \frac{1}{N}\sum_{j=1\atop j\neq i}^Nc_j(\bm x^{(i)})\Big)\\
=~&-\frac{N}{N-1}\Big(c_i(\bm x^{(i)}) - \frac{1}{N}\sum_{j=1}^N c_j(\bm x^{(i)})\Big)\\
=~&-\frac{N}{N-1}\Big[\log\exp\big(c_i(\bm x^{(i)})\big)  - \log\exp\Big(\frac{1}{N}\sum_{j=1}^Nc_j(\bm x^{(i)})\Big)\Big]\\
\nonumber
\leq~&-\frac{N}{N-1}\Big[\log\exp\big(c_i(\bm x^{(i)})\big) \\
    &\qquad\qquad - \log\Big(\frac{1}{N}\sum_{j=1}^N\exp\big(c_j(\bm x^{(i)})\big)\Big)\Big]\\
=~&- \frac{N}{N-1}\Big(\log\frac{\exp\big(c_i(\bm x^{(i)})\big)}{\sum_{j=1}^N\exp\big(c_j(\bm x^{(i)})\big)}+\log N\Big)\\
\label{eq_soft_euc_loss_supp}
=~&- \frac{N}{N-1}\log\frac{\exp\big(c_i(\bm x^{(i)})\big)}{\sum_{j=1}^N\exp\big(c_j(\bm x^{(i)})\big)} - \frac{N\log N}{N-1},
\end{align}
\begin{align}
\nonumber
&\mathcal L_{\text{naive}}(\bm X^{(i)}) \\
=~&-c_i(\bm x^{(i)}) + \frac{1}{N-1}\sum_{j=1\atop j\neq i}^N c_j(\bm x^{(i)})\\
=~&c_i(\bm x^{(i)})+\frac{1}{N-1}\sum_{j=1\atop j\neq i}^N\big(c_j(\bm x^{(i)})\big) - 2c_i(\bm x^{(i)})\\
=~&2\Big[\frac{1}{2}\Big(c_i(\bm x^{(i)})+\frac{1}{N-1}\sum_{j=1\atop j\neq i}^Nc_j(\bm x^{(i)})\Big) - c_i(\bm x^{(i)})\Big]\\
\nonumber
=~&2\Big\{\log\exp\Big[\frac{1}{2}\Big(c_i(\bm x^{(i)}) +\frac{1}{N-1}\sum_{j=1\atop j\neq i}^Nc_j(\bm x^{(i)})\Big)\Big]\\
    &\qquad- \log\exp\big(c_i(\bm x^{(i)})\big)\Big\}\\
\nonumber
\leq~&2\Big\{\log\Big[\frac{1}{2}\Big(\exp\big(c_i(\bm x^{(i)})\big)+\exp\sum_{j=1\atop j\neq i}^N\frac{\big(c_j(\bm x^{(i)})\big)}{N-1}\Big)\Big]\\
    &\qquad- \log\exp\big(c_i(\bm x^{(i)})\big)\Big\}\\
\nonumber
=~&2\Big[\log\Big(\exp\big(c_i(\bm x^{(i)})\big) +\exp\sum_{j=1\atop j\neq i}^N\frac{\big(c_j(\bm x^{(i)})\big)}{N-1}\Big)\\
    &\qquad- \log\exp\big(c_i(\bm x^{(i)})\big)-\log2\Big]\\
\label{eq_loss_positive_term_supp}
=~&2\log\Big(1+\frac{\exp\big(\sum_{j=1\atop j\neq i}^N\frac{c_j(\bm x^{(i)})}{N-1}\big)}{\exp\big(c_i(\bm x^{(i)})\big)}\Big) - 2\log2,
\end{align}
\begin{align}
\nonumber
&\mathcal L_{\text{naive}}(\bm X^{(i)}) \\
=~&-c_i(\bm x^{(i)}) + \frac{1}{N-1}\sum_{j=1\atop j\neq i}^N c_j(\bm x^{(i)})\\
=~&\frac{1}{N-1}\Big(\sum_{j=1\atop j\neq i}^N\big(c_j(\bm x^{(i)})+c_i(\bm x^{(i)})\big) - 2(N-1)c_i(\bm x^{(i)})\Big)\\
=~&\frac{2}{N-1}\Big(\sum_{j=1\atop j\neq i}^N\frac{c_j(\bm x^{(i)})+c_i(\bm x^{(i)})}{2} - (N-1)c_i(\bm x^{(i)})\Big)\\
\nonumber
=~&\frac{2}{N-1}\Big[\sum_{j=1\atop j\neq i}^N\log\exp\frac{c_j(\bm x^{(i)})+c_i(\bm x^{(i)})}{2} \\
        &\qquad\qquad\qquad - (N-1)\log\exp\big(c_i(\bm x^{(i)})\big)\Big]\\
\nonumber
\leq~&\frac{2}{N-1}\Big[\sum_{j=1\atop j\neq i}^N\log\frac{\exp(c_j(\bm x^{(i)}))+\exp(c_i(\bm x^{(i)}))}{2} \\
        &\qquad\qquad\qquad - (N-1)\log\exp\big(c_i(\bm x^{(i)})\big)\Big]\\
\nonumber
=~&\frac{2}{N-1}\sum_{j=1\atop j\neq i}^N\Big[\log\frac{\exp(c_j(\bm x^{(i)}))+\exp(c_i(\bm x^{(i)}))}{2} \\
        &\qquad\qquad\qquad - \log\exp(c_i(\bm x^{(i)}))\Big]\\
=~&\frac{2}{N-1} \sum_{j=1\atop j\neq i}^N\log\frac{\exp(c_j(\bm x^{(i)}))+\exp(c_i(\bm x^{(i)}))}{2\exp(c_i(\bm x^{(i)}))}\\
=~&\frac{2}{N-1} \sum_{j=1\atop j\neq i}^N\log\Big[\frac{1}{2}\Big(1+\frac{\exp(c_j(\bm x^{(i)}))}{\exp(c_i(\bm x^{(i)}))}\Big)\Big]\\
\label{eq_loss_negative_term_supp}
=~&\frac{2}{N-1} \sum_{j=1\atop j\neq i}^N\log\Big(1+\frac{\exp(c_j(\bm x^{(i)}))}{\exp(c_i(\bm x^{(i)}))}\Big)-2\log 2,
\end{align}
and
\begin{align}
\nonumber
&\sum_{i=1}^N \mathcal L_{\text{naive}}(\bm X^{(i)}) \\
=~&\frac{1}{N-1}\Big(\sum_{i=1}^N\sum_{j=1\atop j\neq i}^Nc_j(\bm x^{(i)}) - (N-1)\sum_{i=1}^Nc_i(\bm x^{(i)})\Big)\\
=~&\frac{1}{N-1}\Big(\sum_{i=1}^N\sum_{j=1\atop j\neq i}^Nc_j(\bm x^{(i)}) - \sum_{i=1}^N\sum_{j=1\atop j\neq i}^Nc_j(\bm x^{(j)})\Big)\\
=~&\frac{1}{N-1}\sum_{i=1}^N\sum_{j=1\atop j\neq i}^N\Big(c_j(\bm x^{(i)})  - c_j(\bm x^{(j)})\Big)\\
=~&\frac{1}{N-1}\sum_{i=1}^N\sum_{j=1\atop j\neq i}^N\Big(c_j(\bm x^{(i)}) + c_j(\bm x^{(j)}) - 2c_j(\bm x^{(j)})\Big)\\
 \nonumber
=~&\frac{2}{N-1}\sum_{i=1}^N\sum_{j=1\atop j\neq i}^N\Big(\log\exp\frac{c_j(\bm x^{(i)}) + c_j(\bm x^{(j)})}{2} \\
 &\qquad\qquad - \log\exp\big(c_j(\bm x^{(j)})\big)\Big)\\
 \nonumber
\leq~&\frac{2}{N-1}\sum_{i=1}^N\sum_{j=1\atop j\neq i}^N\Big(\log\frac{\exp(c_j(\bm x^{(i)})) + \exp(c_j(\bm x^{(j)}))}{2} \\
 &\qquad\qquad - \log\exp(c_j(\bm x^{(j)}))\Big)\\
 \nonumber
=~&\frac{2}{N-1}\sum_{i=1}^N\sum_{j=1\atop j\neq i}^N\Big[\log\Big(\exp(c_j(\bm x^{(i)})) + \exp(c_j(\bm x^{(j)}))\Big) \\
 &\qquad\qquad - \log\exp(c_j(\bm x^{(j)}))\Big] - 2N\log2\\
\label{eq_loss_negative_term_sum_supp}
=~&\frac{2}{N-1}\sum_{i=1}^N\sum_{j=1\atop j\neq i}^N\log\Big(1+\frac{\exp(c_j(\bm x^{(i)}))}{\exp(c_j(\bm x^{(j)}))}\Big) - 2N\log2.
\end{align}

\begin{table}[!tbph]
  \centering
  \caption{The results of ResNet50 trained by $L_{\text{soft-nu}}$ with varying $\gamma$ on ImageNet-1K.}\label{Tab_gamma_resnet50-soft-nu_softmax}
  \setlength{\tabcolsep}{0.80mm}{
  \begin{tabular}{c|c ccc cccc c}\hline

$\gamma$                    &1        &2       &3        &4        &5      &6      &7      &8      \\\hline
$\mathcal A_{\text{SW}}$    &49.00    &54.91    &61.21    &65.81    &68.89    &71.68    &73.48    &74.60    \\
$\mathcal A_{\text{CW}}$    &35.40    &41.72    &49.10    &55.02    &59.97    &64.93    &68.04    &\textbf{69.75}    \\
$\mathcal A_{\text{Uni}}$   &18.55    &25.21    &31.67    &39.01    &46.59    &55.26    &60.55    &\textbf{63.47}    \\
$t^*$                       &$-$8.16   &$-$6.93   &$-$5.68   &$-$4.28   &$-$3.34   &4.97   &5.71   &$-$1.45      \\\hline
$\gamma$                    &12     &16     &24     &32     &48     &64     &80     &96\\\hline
$\mathcal A_{\text{SW}}$    &76.07    &76.78    &77.16    &77.21    &\textbf{77.40}    &77.30    &77.18    &77.22    \\
$\mathcal A_{\text{CW}}$    &68.68    &65.64    &61.32    &57.96    &54.20    &52.02    &50.43    &49.53    \\
$\mathcal A_{\text{Uni}}$   &61.85    &57.73    &53.82    &49.49    &44.52    &42.52    &40.97    &39.53    \\
$t^*$                   &6.16   &0.67   &6.58   &4.70   &4.88   &1.64   &2.85   &4.07       \\\hline\hline
$\gamma$                    &112    &128    &114    &160    &176    &192    &208    &256    &512\\\hline
$\mathcal A_{\text{SW}}$    &77.08    &77.13    &77.09    &77.06    &76.93    &76.94    &76.90    &76.86    &76.78\\
$\mathcal A_{\text{CW}}$ &48.60    &47.95    &46.98    &45.73    &45.37    &45.17    &44.90    &44.62    &44.20\\
$\mathcal A_{\text{Uni}}$ &38.23    &37.33    &37.00    &35.87    &35.02    &34.74    &34.67    &34.12    &33.18\\
$t^*$                   &9.40   &9.71   &5.81   &9.78   &9.99   &5.98   &6.50   &6.78   &7.00\\\hline\hline
  \end{tabular}}
\end{table}

\section{The Scalar Factor}

Table \ref{Tab_gamma_resnet50-soft-nu_softmax} presents the results of $L_{\text{soft-nu}}$ on ImageNet-1K, with various $\gamma$.
Although $L_{\text{soft-nu}}$ demonstrates stability with varying $\gamma$,
its uniform accuracy $\mathcal A_{\text{CW}}$ and $\mathcal A_{\text{Uni}}$ are significantly inferior to that of $L_{\text{bce-nu}}$ (Table \ref{Tab_gamma_resnet50-bce-nu}).

\section{SoftMax Function}

For a sequence $\texttt{x} = \{x_i\}_{i=1}^N$, SoftMax function tune-up its each value,
\begin{align}
    \label{eq_softmax_function_supp}
    \text{SoftMax}(\texttt{x})_i = \frac{\exp(x_i)}{\sum_{j=1}^N \exp(x_j)}.
\end{align}
Although SoftMax function does not alter the ordinal relationship among the points in the sequence,
it compresses the gaps of points near the maximum and minimum values,
while stretching the gaps of intermediate points.
Therefore, when we apply SoftMax function on the classification metric sequence $\{c_i(\bm x)\}_{i=1}^N$ of each sample $\bm X$ in the dataset $\mathcal D$,
it does not change the sample-wise classification accuracy $\mathcal A_{\text{SW}}$
but does change the (class-wise) uniform classification accuracy $\mathcal A_{\text{CW}}$ and $\mathcal A_{\text{Uni}}$.

We here present the accuracy of the three models on the six datasets,
by applying the SoftMax function on each sample's classification metric sequence,
as Tables \ref{Tab_imagenet1k_1_softmax} - \ref{Tab_Stanford_Dogs_softmax} show.
Comparing the results in these tables with that in Tables \ref{Tab_CUB_200_2011} - \ref{Tab_Stanford_Dogs},
one can find that the sample-wise accuracy $\mathcal A_{\text{SW}}$ remains unchanged,
while the all uniform accuracies $\mathcal A_{\text{CW}}$ and $\mathcal A_{\text{Uni}}$
show significant improvements.

\begin{table*}[t]
  \centering
  \caption{The classification performances of twelve loss functions on ImageNet-1K, by applying SoftMax function on each sample's classification metric sequence}\label{Tab_imagenet1k_1_softmax}
  \setlength{\tabcolsep}{1.00mm}{
  \begin{tabular}{c|c||c| ccc| ccc| ccc| ccc}\hline
  \multirow{3}{*}{Model} & \multirow{3}{*}{Accuracy} &pre-training&\multicolumn{12}{c}{fine-tuning}\\\cline{3-15}
  &        &$L_{\text{soft-d}}$    &$L_{\text{soft-0}}$    &\makecell[c]{$L_{\text{soft-d}}$\\(baseline)}    &$L_{\text{soft-u}}$    &$L_{\text{soft-n0}}$    &$L_{\text{soft-nd}}$    &$L_{\text{soft-nu}}$
                                    &\color{black}$L_{\text{bce-0}}$    &\color{black}$L_{\text{bce-d}}$    &\color{black}$L_{\text{bce-u}}$    &$L_{\text{bce-n0}}$    &$L_{\text{bce-nd}}$    &$L_{\text{bce-nu}}$\\\hline\hline
  \multirow{4}{*}{ResNet50}&$\mathcal A_{\text{SW}}$  &75.80  &76.82    &76.74    &76.87    &77.21    &77.07    &77.22    &77.00    &77.12    &76.99    &77.34    &77.38    &\textbf{77.50}\\
&$\mathcal A_{\text{CW}}$  &73.94&75.08    &75.07    &75.03    &75.62    &75.50    &75.63    &75.66    &75.72    &75.73    &76.24    &76.28    &\textbf{76.46}\\
&$\mathcal A_{\text{Uni}}$ &72.00&73.23    &73.29    &73.24    &73.97    &73.77    &73.93    &74.21    &74.22    &74.30    &74.91    &74.88    &\textbf{75.07}\\
&$t^*$      &0.388&0.384   &0.412   &0.386   &0.407   &0.392   &0.412   &0.379   &0.382   &0.381   &0.413   &0.379   &0.395\\\hline
  \multirow{4}{*}{ResNet101}&$\mathcal A_{\text{SW}}$  &77.26   &78.39    &78.47    &78.42    &78.62    &78.72    &78.59    &78.92    &78.88    &79.01    &79.11    &79.06    &\textbf{79.16}\\
&$\mathcal A_{\text{CW}}$  &75.77&76.97    &76.99    &77.00    &77.41    &77.44    &77.31    &77.81    &77.84    &78.01    &78.24    &78.23    &\textbf{78.38}\\
&$\mathcal A_{\text{Uni}}$ &74.04&75.34    &75.25    &75.29    &75.86    &75.98    &75.79    &76.51    &76.52    &76.69    &77.08    &77.12    &\textbf{77.16}\\
&$t^*$      &0.418&0.405   &0.399   &0.391   &0.395   &0.433   &0.405   &0.423   &0.384  &0.408   &0.414   &0.418   &0.409\\\hline
  \multirow{4}{*}{DenseNet161}&$\mathcal A_{\text{SW}}$  &77.38&78.51    &78.58    &78.64    &78.05    &78.05    &77.94    &79.07    &79.19    &79.15    &79.24    &79.10    &\textbf{79.29}\\
&$\mathcal A_{\text{CW}}$  &76.04&77.27    &77.23    &77.28    &76.89    &76.88    &76.81    &78.28    &78.35    &78.33    &78.29    &78.30    &\textbf{78.46}\\
&$\mathcal A_{\text{Uni}}$ &74.90&75.68    &75.62    &75.78    &75.43    &75.43    &75.32    &77.05    &77.11    &77.12    &77.07    &77.12    &\textbf{77.26}\\
&$t^*$      &0.407&0.405   &0.412   &0.425   &0.418   &0.418   &0.415   &0.417   &0.396   &0.386   &0.409   &0.389   &0.402\\\hline

  \end{tabular}}
\end{table*}

\begin{table*}[!tbph]
  \centering
  \caption{The classification performances of twelve loss functions on CUB, by applying SoftMax function on each sample's classification metric sequence}\label{Tab_CUB_200_2011_softmax}
  \setlength{\tabcolsep}{1.00mm}{
  \begin{tabular}{c|c|| ccc| ccc| ccc| ccc}\hline
  Model & Accuracy &$L_{\text{soft-0}}$    &\makecell[c]{$L_{\text{soft-d}}$\\(baseline)}    &$L_{\text{soft-u}}$    &$L_{\text{soft-n0}}$    &$L_{\text{soft-nd}}$    &$L_{\text{soft-nu}}$
                                    &$L_{\text{bce-0}}$    &$L_{\text{bce-d}}$    &$L_{\text{bce-u}}$    &$L_{\text{bce-n0}}$    &$L_{\text{bce-nd}}$    &$L_{\text{bce-nu}}$\\\hline\hline

\multirow{4}{*}{ResNet50} &$\mathcal A_{\text{SW}}$  &81.98    &81.86    &81.84    &84.88    &84.93    &84.93    &79.46    &79.39    &79.27    &84.00    &85.48    &\textbf{86.09}\\
&$\mathcal A_{\text{CW}}$  &81.60    &81.60    &81.46    &84.71    &84.76    &84.71    &78.82    &78.96    &78.63    &83.78    &85.43    &\textbf{86.07}\\
&$\mathcal A_{\text{Uni}}$ &80.69    &80.76    &80.60    &84.02    &84.17    &84.17    &77.51    &77.74    &77.65    &83.15    &85.11    &\textbf{85.69}\\
&$t^*$      &0.443    &0.441    &0.434    &0.461    &0.458    &0.472    &0.429   &0.397   &0.385   &0.456    &0.473    &0.483\\\hline

\multirow{4}{*}{ResNet101} &$\mathcal A_{\text{SW}}$  &82.97    &83.24    &83.48    &86.33    &85.57    &85.57    &79.44    &79.82    &80.45    &85.85    &\textbf{86.83}    &86.28\\
&$\mathcal A_{\text{CW}}$  &82.72    &82.93    &83.19    &86.18    &85.42    &85.45    &79.24    &79.74    &80.34    &85.80    &\textbf{86.80}    &86.21\\
&$\mathcal A_{\text{Uni}}$ &82.17    &82.17    &82.45    &85.47    &85.00    &84.98    &78.56    &79.24    &79.94    &85.57    &\textbf{86.47}    &85.97\\
&$t^*$      &0.469   &0.442   &0.447   &0.465   &0.454   &0.435   &0.438   &0.469   &0.469   &0.488   &0.439   &0.476\\\hline

\multirow{4}{*}{DenseNet161} &$\mathcal A_{\text{SW}}$&83.79    &83.85    &84.04    &86.97    &86.78    &86.37    &77.93    &77.70    &77.99    &\textbf{87.47}    &87.11    &87.25\\
&$\mathcal A_{\text{CW}}$  &83.36    &83.53    &83.76    &86.62    &86.64    &86.05    &77.74    &77.53    &77.79    &\textbf{87.33}    &87.04    &87.11\\
&$\mathcal A_{\text{Uni}}$ &82.59    &82.67    &82.88    &85.86    &85.90    &84.97    &77.25    &76.91    &77.24    &\textbf{86.81}    &86.68    &86.68\\
&$t^*$      &0.467   &0.464   &0.403   &0.450   &0.474   &0.404   &0.445   &0.461   &0.436   &0.454   &0.495   &0.461\\\hline
  \end{tabular}}
\end{table*}
\begin{table*}[!tbph]
  \centering
  \caption{The classification performances of twelve loss functions on Aircraft, by applying SoftMax function on each sample's classification metric sequence}\label{Tab_FGVC-aircraft_softmax}
  \setlength{\tabcolsep}{1.00mm}{
  \begin{tabular}{c|c|| ccc| ccc| ccc| ccc}\hline
  Model & Accuracy &$L_{\text{soft-0}}$    &\makecell[c]{$L_{\text{soft-d}}$\\(baseline)}    &$L_{\text{soft-u}}$    &$L_{\text{soft-n0}}$    &$L_{\text{soft-nd}}$    &$L_{\text{soft-nu}}$
                                    &$L_{\text{bce-0}}$    &$L_{\text{bce-d}}$    &$L_{\text{bce-u}}$    &$L_{\text{bce-n0}}$    &$L_{\text{bce-nd}}$    &$L_{\text{bce-nu}}$\\\hline\hline

\multirow{4}{*}{ResNet50} &$\mathcal A_{\text{SW}}$  &91.42    &91.48    &91.39    &91.33    &91.24    &91.39    &90.52    &90.46    &90.73    &90.82    &92.11    &\textbf{92.14}\\
&$\mathcal A_{\text{CW}}$  &91.33    &91.33    &91.27    &91.27    &91.21    &91.33    &90.43    &90.46    &90.67    &90.79    &92.05    &\textbf{92.08}\\
&$\mathcal A_{\text{Uni}}$ &91.12    &90.97    &90.91    &90.82    &90.94    &90.91    &90.13    &90.22    &90.37    &90.46    &91.93    &\textbf{91.93}\\
&$t^*$      &0.477   &0.499   &0.495   &0.497   &0.462   &0.471   &0.403   &0.456  &0.485   &0.475   &0.494   &0.479\\\hline

\multirow{4}{*}{ResNet101} &$\mathcal A_{\text{SW}}$  &92.11    &91.87    &92.05    &91.90    &92.05    &91.87    &90.82    &91.06    &90.85    &91.51    &92.08    &\textbf{92.41}\\
&$\mathcal A_{\text{CW}}$  &92.11    &91.87    &92.02    &91.78    &92.05    &91.81    &90.79    &91.00    &90.82    &91.51    &91.99    &\textbf{92.41}\\
&$\mathcal A_{\text{Uni}}$ &92.05    &91.75    &91.87    &91.66    &91.78    &91.63    &90.55    &90.82    &90.61    &91.33    &91.93    &\textbf{92.29}\\
&$t^*$      &0.498   &0.487   &0.450   &0.480   &0.485   &0.495   &0.418   &0.434   &0.488   &0.403   &0.462   &0.379\\\hline

\multirow{4}{*}{DenseNet161} &$\mathcal A_{\text{SW}}$ &91.99    &92.20    &\textbf{92.71}    &91.90    &91.21    &91.63    &91.33    &91.24    &91.36    &92.44    &92.44    &92.38\\
&$\mathcal A_{\text{CW}}$  &91.84    &92.05    &\textbf{92.71}    &91.84    &91.06    &91.51    &91.30    &91.24    &91.33    &92.44    &92.41    &92.32\\
&$\mathcal A_{\text{Uni}}$ &91.63    &91.90    &\textbf{92.35}    &91.42    &90.70    &91.18    &91.12    &91.00    &91.21    &92.29    &92.05    &92.14\\
&$t^*$      &0.441   &0.499   &0.459   &0.422   &0.488   &0.453   &0.454   &0.451   &0.389   &0.474   &0.491   &0.454\\\hline
  \end{tabular}}
\end{table*}
\begin{table*}[!tbph]
  \centering
  \caption{The classification performances of twelve loss functions on Flowers102, by applying SoftMax function on each sample's classification metric sequence}\label{Tab_Flowers102_softmax}
  \setlength{\tabcolsep}{1.00mm}{
  \begin{tabular}{c|c|| ccc| ccc| ccc| ccc}\hline
  Model & Accuracy &$L_{\text{soft-0}}$    &\makecell[c]{$L_{\text{soft-d}}$\\(baseline)}    &$L_{\text{soft-u}}$    &$L_{\text{soft-n0}}$    &$L_{\text{soft-nd}}$    &$L_{\text{soft-nu}}$
                                    &$L_{\text{bce-0}}$    &$L_{\text{bce-d}}$    &$L_{\text{bce-u}}$    &$L_{\text{bce-n0}}$    &$L_{\text{bce-nd}}$    &$L_{\text{bce-nu}}$\\\hline\hline

\multirow{4}{*}{ResNet50} &$\mathcal A_{\text{SW}}$ &95.33    &95.97    &95.90    &96.76    &96.94    &96.28    &95.58    &95.51    &95.41    &\textbf{97.20}    &96.91    &97.15\\
&$\mathcal A_{\text{CW}}$  &95.17    &95.71    &95.67    &96.67    &96.86    &96.13    &95.30    &95.32    &94.99    &\textbf{97.17}    &96.89    &97.14\\
&$\mathcal A_{\text{Uni}}$ &94.84    &95.53    &95.37    &96.52    &96.68    &95.95    &94.91    &94.91    &94.57    &\textbf{97.01}    &96.88    &97.01\\
&$t^*$       &0.432   &0.428   &0.448   &0.449   &0.475   &0.430   &0.377   &0.409   &0.383   &0.495   &0.476   &0.448\\\hline

\multirow{4}{*}{ResNet101} &$\mathcal A_{\text{SW}}$  &96.23    &96.52    &96.71    &96.93    &97.27    &97.01    &95.87    &96.06    &96.26    &97.67    &\textbf{97.79}    &97.66\\
&$\mathcal A_{\text{CW}}$  &96.02    &96.37    &96.63    &96.83    &97.24    &96.93    &95.76    &96.02    &96.13    &97.61    &\textbf{97.76}    &97.59\\
&$\mathcal A_{\text{Uni}}$ &95.82    &96.10    &96.39    &96.65    &97.06    &96.68    &95.56    &95.85    &95.79    &97.40    &\textbf{97.67}    &97.51\\
&$t^*$       &0.398   &0.420   &0.447   &0.432   &0.423   &0.392   &0.388   &0.465   &0.413   &0.493   &0.471   &0.457\\\hline

\multirow{4}{*}{DenseNet161} &$\mathcal A_{\text{SW}}$ &96.96    &96.58    &96.81    &96.76    &96.73    &96.80    &96.85    &97.09    &96.08    &\textbf{98.00}    &97.76    &97.69\\
&$\mathcal A_{\text{CW}}$  &96.70    &96.26    &96.60    &96.41    &96.47    &96.49    &96.71    &97.01    &96.67    &\textbf{97.85}    &97.61    &97.54\\
&$\mathcal A_{\text{Uni}}$ &96.36    &95.92    &96.42    &96.06    &96.16    &96.21    &96.55    &96.88    &96.55    &\textbf{97.58}    &97.43    &97.40\\
&$t^*$       &0.355   &0.356   &0.369   &0.359   &0.371   &0.378   &0.420   &0.406   &0.328   &0.336   &0.373   &0.400\\\hline
  \end{tabular}}
\end{table*}
\begin{table*}[!tbph]
  \centering
  \caption{The classification performances of twelve loss functions on Cars, by applying SoftMax function on each sample's classification metric sequence}\label{Tab_Stanford_Cars_softmax}
  \setlength{\tabcolsep}{1.00mm}{
  \begin{tabular}{c|c|| ccc| ccc| ccc| ccc}\hline
  Model & Accuracy &$L_{\text{soft-0}}$    &\makecell[c]{$L_{\text{soft-d}}$\\(baseline)}    &$L_{\text{soft-u}}$    &$L_{\text{soft-n0}}$    &$L_{\text{soft-nd}}$    &$L_{\text{soft-nu}}$
                                    &$L_{\text{bce-0}}$    &$L_{\text{bce-d}}$    &$L_{\text{bce-u}}$    &$L_{\text{bce-n0}}$    &$L_{\text{bce-nd}}$    &$L_{\text{bce-nu}}$\\\hline\hline

\multirow{4}{*}{ResNet50} &$\mathcal A_{\text{SW}}$ &93.21    &93.22    &93.16    &93.68    &93.36    &93.28    &92.87    &92.59    &92.44    &92.65    &\textbf{93.91}    &93.86\\
&$\mathcal A_{\text{CW}}$ &93.17    &93.14    &93.12    &93.67    &93.31    &93.17    &92.82    &92.49    &92.38    &92.56    &\textbf{93.88}    &93.84\\
&$\mathcal A_{\text{Uni}}$ &92.84    &92.71    &92.80    &93.37    &93.06    &92.86    &92.44    &92.21    &91.93    &92.21    &\textbf{93.74}    &93.67\\
&$t^*$      &0.462   &0.486   &0.486   &0.470   &0.469   &0.479   &0.449   &0.497   &0.467   &0.412   &0.439   &0.456\\\hline

\multirow{4}{*}{ResNet101} &$\mathcal A_{\text{SW}}$ &93.30    &93.51    &93.47    &93.57    &93.53    &93.62    &92.76    &92.30    &92.58    &93.69    &\textbf{94.23}    &94.14\\
&$\mathcal A_{\text{CW}}$  &93.26    &93.48    &93.42    &93.51    &93.52    &93.62    &92.74    &92.28    &92.56    &93.69    &\textbf{94.20}    &94.13\\
&$\mathcal A_{\text{Uni}}$ &92.99    &93.20    &93.14    &93.25    &93.22    &93.32    &92.51    &92.07    &92.39    &93.55    &\textbf{94.09}    &93.94\\
&$t^*$      &0.448   &0.487   &0.497   &0.450   &0.465   &0.482   &0.450   &0.425   &0.480   &0.483   &0.497   &0.496\\\hline

\multirow{4}{*}{DenseNet161} &$\mathcal A_{\text{SW}}$ &92.96    &92.96    &93.32    &92.55    &92.35    &92.75    &91.79    &91.46    &91.43    &\textbf{93.63}    &93.22    &93.53\\
&$\mathcal A_{\text{CW}}$  &92.91    &92.91    &93.25    &92.35    &92.23    &92.61    &91.78    &91.39    &91.39    &\textbf{93.57}    &93.17    &93.50\\
&$\mathcal A_{\text{Uni}}$ &92.59    &92.50    &92.85    &91.74    &91.57    &92.05    &91.56    &91.16    &91.16    &\textbf{93.41}    &92.96    &93.25\\
&$t^*$      &0.478   &0.477   &0.468   &0.407   &0.412   &0.460   &0.452   &0.478   &0.479   &0.460   &0.478   &0.445\\\hline
  \end{tabular}}
\end{table*}
\begin{table*}[!tbph]
  \centering
  \caption{The classification performances of twelve loss functions on Dogs, by applying SoftMax function on each sample's classification metric sequence}\label{Tab_Stanford_Dogs_softmax}
  \setlength{\tabcolsep}{1.00mm}{
  \begin{tabular}{c|c|| ccc| ccc| ccc| ccc}\hline
  Model & Accuracy &$L_{\text{soft-0}}$    &\makecell[c]{$L_{\text{soft-d}}$\\(baseline)}    &$L_{\text{soft-u}}$    &$L_{\text{soft-n0}}$    &$L_{\text{soft-nd}}$    &$L_{\text{soft-nu}}$
                                    &$L_{\text{bce-0}}$    &$L_{\text{bce-d}}$    &$L_{\text{bce-u}}$    &$L_{\text{bce-n0}}$    &$L_{\text{bce-nd}}$    &$L_{\text{bce-nu}}$\\\hline\hline

\multirow{4}{*}{ResNet50} &$\mathcal A_{\text{SW}}$  &81.82    &81.83    &81.20    &82.60    &82.70    &82.66    &81.13    &81.26    &81.35    &82.45    &83.11    &\textbf{83.18}\\
&$\mathcal A_{\text{CW}}$  &81.61    &81.56    &80.91    &82.41    &82.38    &82.51    &80.87    &80.99    &81.12    &82.35    &83.02    &\textbf{83.08}\\
&$\mathcal A_{\text{Uni}}$ &81.13    &81.17    &80.44    &82.06    &82.02    &82.21    &80.48    &80.54    &80.82    &82.02    &\textbf{82.89}    &82.86\\
&$t^*$      &0.487   &0.479   &0.471   &0.471   &0.478   &0.479   &0.461   &0.458   &0.478   &0.491   &0.490   &0.480\\\hline

\multirow{4}{*}{ResNet101} &$\mathcal A_{\text{SW}}$ &83.86    &83.53    &83.39    &84.23    &84.17    &83.76    &83.24    &82.48    &82.63    &83.28    &\textbf{84.31}    &83.94\\
&$\mathcal A_{\text{CW}}$  &83.71    &83.33    &83.18    &84.06    &83.93    &83.64    &83.10    &82.34    &82.46    &83.24    &\textbf{84.24}    &83.86\\
&$\mathcal A_{\text{Uni}}$ &83.36    &82.97    &82.76    &83.79    &83.50    &83.36    &82.86    &82.16    &82.21    &83.09    &\textbf{84.03}    &83.72\\
&$t^*$      &0.472   &0.483   &0.485  &0.482   &0.486   &0.472   &0.486   &0.485   &0.496   &0.461   &0.493   &0.475\\\hline

\multirow{4}{*}{DenseNet161} &$\mathcal A_{\text{SW}}$ &84.20    &84.53    &84.71    &85.05    &84.92    &84.91    &82.68    &82.67    &82.63    &84.84    &84.92    &\textbf{85.05}\\
&$\mathcal A_{\text{CW}}$  &84.02    &84.34    &84.56    &84.79    &84.72    &84.60    &82.58    &82.49    &82.54    &84.79    &84.50    &\textbf{85.00}\\
&$\mathcal A_{\text{Uni}}$ &83.60    &84.01    &84.24    &84.43    &84.36    &84.07    &82.33    &82.25    &82.33    &84.59    &83.69    &\textbf{84.77}\\
&$t^*$      &0.456   &0.492   &0.443   &0.469   &0.493   &0.477   &0.487   &0.498   &0.471   &0.500   &0.469   &0.487\\\hline
  \end{tabular}}
\end{table*}

\end{document}